\documentclass[journal]{IEEEtran}
\usepackage{amsmath,amsfonts}
\usepackage{algorithmic}
\usepackage{array}
\usepackage[caption=false,font=normalsize,labelfont=sf,textfont=sf]{subfig}
\usepackage{textcomp}
\usepackage{stfloats}
\usepackage{url}
\usepackage{verbatim}
\usepackage{multirow}
\usepackage{graphicx}
\usepackage{wrapfig}
\usepackage{bm}
\usepackage{amsfonts}       
\usepackage{nicefrac}       
\usepackage{xcolor}         %
\usepackage{tabularx} 
\usepackage{ragged2e} 
\usepackage{booktabs}
\usepackage[caption=false]{subfig}
\usepackage{cite}  
\usepackage{microtype}      
\usepackage{balance}

\newcommand{\eg}{\textit{e.g., }}
\newcommand{\ie}{\textit{i.e., }}

\begin{document}
\title{Revisiting Federated Fine-Tuning: \\ A Single Communication Round is Enough \\ for Foundation Models}
\author{Ziyao Wang, Bowei Tian, Yexiao He, Zheyu Shen, Guoheng Sun, Yuhan Liu, Luyang Liu,

Meng Liu, and Ang Li
\thanks{Ziyao Wang, Bowei Tian, Yexiao He, Zheyu Shen, Guoheng Sun, Meng Liu, and Ang Li are with the University of Maryland, College Park, MD, USA. Emails: ziyaow@umd.edu, angliece@umd.edu. Yuhan Liu is with the Queen Mary University of London. Luyang Liu is with Google DeepMind.}}

\markboth{Journal of \LaTeX\ Class Files,~Vol.~18, No.~9, September~2020}%
{How to Use the IEEEtran \LaTeX \ Templates}

\maketitle

\begin{abstract}
The recent advancement of foundation models (FMs) has increased the demand for fine-tuning these models on large-scale cross-domain datasets. To address this, federated fine-tuning has emerged, allowing FMs to be fine-tuned on distributed datasets across multiple devices while ensuring data privacy. 
However, the substantial parameter size and the multi-round communication in federated learning algorithms result in prohibitively high communication costs, challenging the practicality of federated fine-tuning.
In this paper, we identify and analyze, both theoretically and empirically, that the traditional multi-round aggregation algorithms may not be necessary for federated fine-tuning large FMs.
Our experiments reveal that a single round of aggregation (\ie one-shot federated fine-tuning) yields a global model performance comparable to that achieved through multiple rounds of aggregation.
Through rigorous mathematical and empirical analyses, we demonstrate that large FMs, due to their extensive parameter sizes and pre-training on general tasks, achieve significantly lower training loss in one-shot federated fine-tuning compared to smaller models.
Our extensive experiments show that one-shot federated fine-tuning significantly reduces communication costs. It also has the potential to enable asynchronous aggregation, enhances privacy, and maintains performance consistency with multi-round federated fine-tuning on both text generation and text-to-image generation tasks.
Our findings provide insights to revolutionize federated fine-tuning in practice, enhancing efficiency, reducing costs, and expanding accessibility for FMs. 
\end{abstract}


\section{Introduction}
\IEEEPARstart{C}{utting-edge} foundation models (FMs) demonstrate remarkable versatility across various domains. Notably, large language models (LLMs) like GPT-4~\cite{achiam2023gpt}, Gemma~\cite{team2024gemma}, and Llama~\cite{touvron2023llama2} excel in tasks such as translation, question answering (QA), chat assistant, and math. 
Similarly, diffusion models~\cite{song2020score,yang2023diffusion} can generate diverse images based on textual descriptions. Achieving such versatility requires fine-tuning these FMs on cross-domain datasets.  However, this process faces significant challenges in real-world scenarios due to the valuable datasets residing on devices owned by organizations or individuals, raising privacy concerns. 
To address these privacy issues, researchers have proposed using federated learning (FL)~\cite{mcmahan2017communication,li2020lotteryfl,li2021hermes} for distributed fine-tuning of FMs, a process known as federated fine-tuning~\cite{zhang2024towards,ye2024openfedllm,yao2409federated}. Federated fine-tuning allows distributed clients to collaboratively fine-tune a global FM on specific tasks without disclosing their private data.

Traditional FL requires \emph{multiple communication rounds} between clients and the server to ensure the global model convergence~\cite{mcmahan2017communication}. However, the substantial parameter size of FMs (typically in billions) results in significant communication overhead~\cite{ghiasvand2024communication}. Many devices lack the capability to repeatedly communicate model parameters of this scale~\cite{liu2022chronos}. While previous works adopt parameter-efficient fine-tuning (PEFT) methods such as low-rank adaptation (LoRA)~\cite{hu2021lora,yang2024low,shen2025edgelora, wang2025prada} to reduce the number of trainable and communicated parameters, the high communication requirements of federated fine-tuning remain a practical limitation~\cite{sun2024improving,wang2024flora,yi2023pfedlora,wang2025fedmabench}.

\begin{figure}[t]
  \centering
  \includegraphics[width=0.5\textwidth]{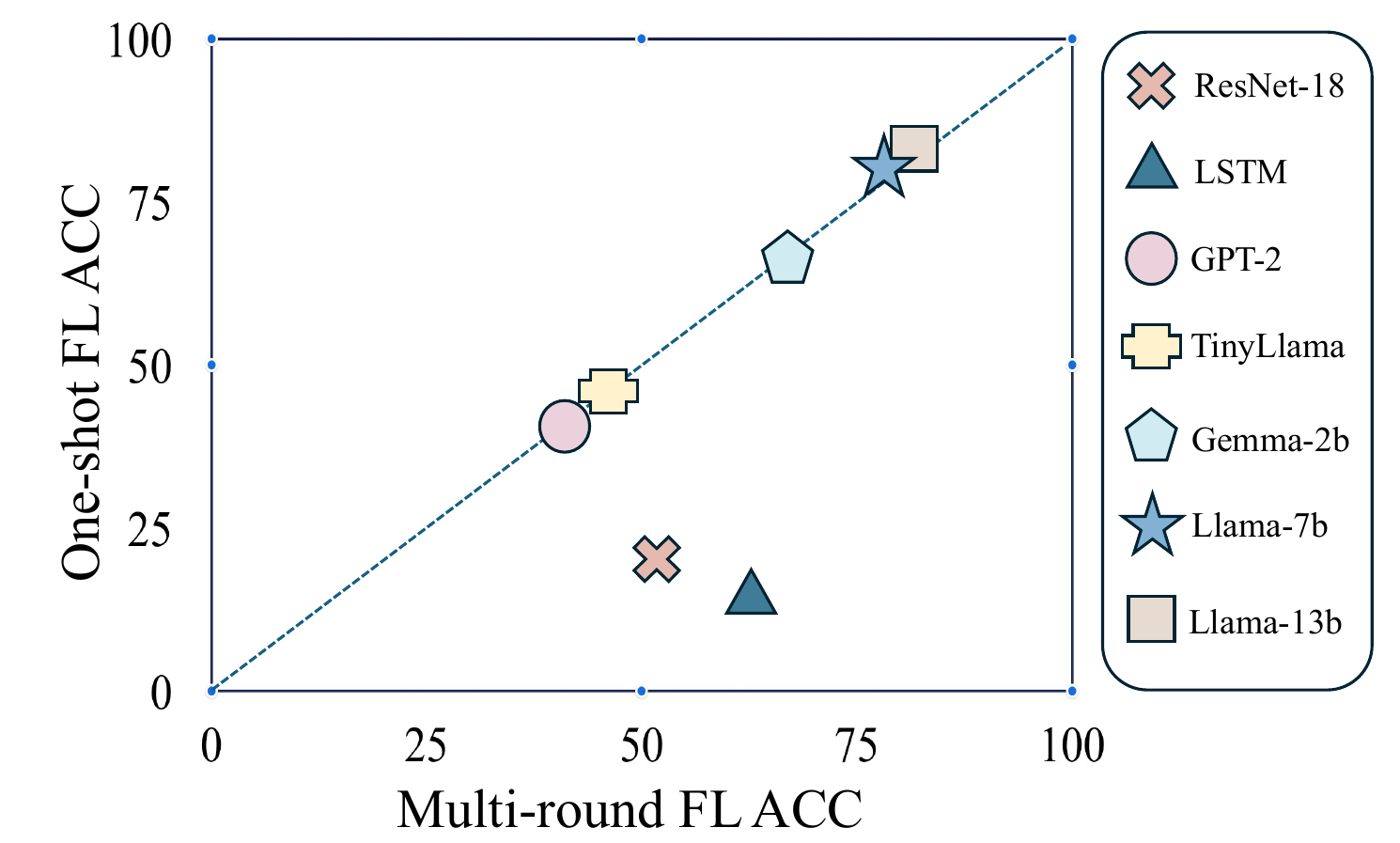}
  \caption{The distinct performances of one-shot federated learning between small models and large FMs. The horizontal axis represents multi-round FL accuracy, while the vertical axis represents one-shot FL accuracy. The ResNet-18 and LSTM are trained and tested on CIFAR-10 and Shakespeare respectively. Other models are fine-tuned on Wizard dataset and tested on ARC Easy. The closer the points are to the dashed line means the accuracy of one-shot and multi-round FL are closer in the corresponding model.}
  \label{fig:compare}
\end{figure}

Unexpectedly, our recent experiments have discovered an emergent capability of FMs that could fundamentally shift the approach to federated fine-tuning. We find that with sufficient local fine-tuning epochs, \textbf{\textit{a single communication round is enough to effectively fine-tune FMs}}, which is called \emph{one-shot federated fine-tuning}~\cite{guha2019one}. 
Fig. \ref{fig:compare} highlights the performance comparisons between one-shot FL and traditional multi-round FL, maintaining the same total number of local epochs. While one-shot FL underperforms multi-round FL for smaller models (\eg ResNet-18 and LSTM), it achieves comparable performance for larger FMs (\eg GPT-2, Llama, etc). 
This unique discovery challenges the conventional belief that multiple communication rounds are essential for the federated fine-tuning of FMs. Instead, we demonstrate that FMs can achieve convergence with just a single aggregation of well-fine-tuned local models.
This paper explores this innovative finding, providing rigorous theoretical analysis and compelling empirical evidence to validate the effectiveness of one-shot federated fine-tuning FMs. Our theory for why one-shot FL works well in larger models also sheds light on the success of task vectors~\cite{hendel2023context,wang2024moderator}, task arithmetic~\cite{ilharco2022editing}, and model merging~\cite{zhao2024texttt} in FMs.

Beyond the theoretical contributions, an effective one-shot FL also offers advantages for the practical deployment of FL systems.
\underline{First}, it \textbf{dramatically reduces communication costs}. One-Shot FL slashes communication overhead by a factor of $\frac{1}{T}$, where $T$ represents the number of communication rounds in traditional federated fine-tuning. This reduction is a game-changer for devices with limited bandwidth.
\underline{Second}, one-shot FL enables seamless \textbf{asynchronous training}. This flexibility removes the bottleneck of server waiting times, ensuring uninterrupted training regardless of client connectivity or resource limitations. The process becomes far more robust and efficient.
\underline{Third}, one-shot FL offers \textbf{enhanced security} against prevalent client-side federated learning attacks. Attacks like client-side model inversion and gradient inversion, which depend on multiple global model updates, 
are rendered ineffective. This significantly bolsters the integrity of the training process.

Our key contributions are listed as follows:
\begin{itemize}
    \item \textbf{Empirical Insight:} We identify and systematically validate the surprising effectiveness of one-shot federated fine-tuning of large foundation models.
    
    \item \textbf{Theoretical Analysis:} We theoretically demonstrate the relationship between the error of one-shot federated fine-tuning and model smoothness, fine-tuning model update, and number of fine-tuning rounds. Our analysis, supported by experiments, reveals that large FMs are smoother, exhibit smaller model updates, and require fewer fine-tuning epochs than smaller models, resulting in significantly lower one-shot federated fine-tuning errors.
    
    \item \textbf{Experimental Validation:} We conduct extensive experiments on six FMs and three tasks, demonstrating that one-shot federated fine-tuning achieves performance comparable to multi-round federated fine-tuning, particularly for models with over 1 billion parameters. Experimental results also surprisingly show that LoRA outperforms full fine-tuning in some cases of one-shot federated fine-tuning.
\end{itemize} 
 
\section{Preliminary}
\textbf{Federated Learning Paradigm of Small Models.}
In FL, the primary objective is to optimize a global objective function $F(\bm{w})$, which is weighted average of the local objective functions from $m$ clients~\cite{wang2020tackling}:
\begin{equation}
F(\bm{w})=\sum_{i=1}^{m}p_i F_i(\bm{w}),
\label{eq:avg_gl}
\end{equation}
where $\bm{w}$ represents the model parameters and $p_i$ is the scaling factor. To protect the data privacy of each client, the server cannot access the local dataset. Thus, the local objective function $F_i(\bm{w})$ remains unknown to the server. FedAvg~\cite{mcmahan2017communication} algorithm provides a distributed training algorithm to facilitate privacy-conscious training. It allows multiple clients to train the model on their local datasets and aggregates locally trained models on the server at the end of each communication round. In the $t$-th communication round, the global model update rule of FedAvg is: 
\begin{equation}
\bm{w}^{(t+1,0)} - \bm{w}^{(t,0)} = \alpha^{(t)}\sum_{i=1}^{m}p_i \Delta_i^{(t)}, \quad t \in [0, T-1],
\label{eq:avg_up}
\end{equation}
where $\bm{w}^{(t,0)}$ is the model weights in $t$-th round and 0-th local epoch, which represents the global model in $t$-th round. $T$ is the total number of communication rounds, $\alpha^{(t)}$ is the global learning rate, and $\Delta_i^{(t)}$ is the local model update in $t$-th round. $\Delta_i^{(t)}$ is the accumulative model update of $k$ local stochastic gradient descent (SGD) steps:
\begin{equation}
\Delta_i^{(t)} = \sum_{j=1}^{k}\beta_i^{(t,j)}g_i(\bm{w}_i^{(t,j)}),
\label{eq:avg_lc}
\end{equation}
where $g_i(\bm{w}_i^{(t,j)})$ is the stochastic gradient over a local mini-batch and $\beta_i^{(t,j)}$ is the local learning rate. Note that $j$ here represents a mini-batch, and $k$ is the total number of mini-batches per client.

Local datasets in FL are typically heterogeneous, leading to differences in local objectives. Therefore, FL usually converges more slowly than centralized machine learning~\cite{mcmahan2017communication,wang2023fedhyper}. This slow convergence necessitates a large number of global communication rounds and local epochs to achieve satisfactory performance.
For example, experimental results in \cite{reddi2020adaptive} show that the ResNet-18 model requires more than 2000 and 4000 communication rounds to converge on CIFAR-10~\cite{krizhevsky2009learning} and CIFAR-100 respectively. Even for simple natural language processing tasks such as Shakespeare, an RNN model needs more than 50 rounds to converge. 
The requirement for multi-round communication rounds introduces several significant drawbacks. 
First, clients must frequently exchange model parameters with the server, which can be prohibitively expensive in certain constrained scenarios or on devices with limited resources. 
Second, repeated invocation of computational resources for training increases the overall computational overhead. 
Additionally, the multi-round communication approach leads to excessive energy consumption, synchronization difficulties, and challenges in maintaining privacy protection. 
Thus, optimizing FL algorithms to minimize the number of communication rounds is an essential research direction in FL.

\textbf{Federated Fine-Tuning Foundation Models.} Foundation models~\cite{zhou2023comprehensive} refer to pre-trained deep learning models with a vast number of parameters, typically in the order of billions. These FMs are trained on broad data at scale and are adaptable to a wide range of downstream tasks when fine-tuned on domain-specific datasets~\cite{bommasani2021opportunities}. Since domain-specific datasets are often distributed across multiple devices, FL offers an important paradigm for fine-tuning FMs while preserving data privacy.

Federated fine-tuning~\cite{sun2024improving,wang2024flora,cheng2021fine,woisetschlager2024federated,hilmkil2021scaling} adopts the same FedAvg algorithm in Eq. \ref{eq:avg_gl} and Eq. \ref{eq:avg_up} to aggregate the local model updates. The key difference lies in the \textit{model parameter size}. 
The parameter size of FMs is usually hundreds of times greater than that of small models, resulting in a significant increase in the computation resources and communication overhead required for federated fine-tuning. 
Given the network communication capabilities of commonly used devices, performing multi-round synchronized communication of large model parameters between servers and clients is virtually impossible. Although parameter-efficient fine-tuning algorithms like LoRA~\cite{hu2021lora} have been adopted in federated fine-tuning, the communication overhead remains excessively high, hindering practical application.

\textbf{One-Shot Federated Learning.} To reduce communication overhead in FL, recent works have focused on one-shot  FL~\cite{jhunjhunwala2024fedfisher,guha2019one,gong2021ensemble,li2020practical,zhou2020distilled,yang2024exploring}, which uses a single communication round to obtain the global model. These algorithms often employ knowledge distillation or neuron-matching methods to optimize the global model.
However, these approaches require additional data or computation. Knowledge distillation often necessitates auxiliary public datasets or external generative models, and neuron matching requires additional computation on both clients and the server. Despite these additional resource requirements, the performance of one-shot FL has historically been inferior to standard multi-round FL. For instance, experiments in \cite{jhunjhunwala2024fedfisher} show that one-shot FL achieves only 50\% accuracy on the CIFAR-10 dataset, which is 20\% lower than the accuracy achieved with 5-round FL.

However, our recent experiments have uncovered greater potential for one-shot federated fine-tuning large FMs. As shown in Fig. \ref{fig:compare}, one-shot FL for large models does not show a significant performance gap compared to multi-round FL, which is commonly observed with smaller models. In fact, when the total number of local epochs is the same, the performance of large models fine-tuned by one-shot FL is comparable to that of multi-round FL. Additionally, in fine-tuning larger models such as Llama-13b, one-shot FL even performed slightly better than multi-round FL.
These results, along with the experiment results in Section \ref{sec:exp}, suggest that traditional multi-round FL algorithms may no longer be necessary for federated fine-tuning large FMs. Large FMs can effectively learn downstream tasks from distributed clients with just a single communication round, opening up new possibilities for federated fine-tuning applications.

Although we have observed consistently good performance with one-shot federated fine-tuning, the reasons behind this phenomenon remain unexplored. In the next section, we will delve into this phenomenon through theoretical analysis.
 
\section{Theoretical Analysis of One-Shot Federated Fine-Tuning}\label{sec:theor}
For a multi-round FL algorithm, if the total number of communication rounds is $T$ and the number of local SGD steps for each round is $k$, according to Eq. \ref{eq:avg_up} the global model parameters after FL satisfy:
\begin{equation}
\bm{w}^{(T,0)} - \bm{w}^{(0,0)} = \sum_{t=0}^{T-1}\alpha^{(t)}\sum_{i=1}^{m}p_i \Delta_i^{(t)},
\label{eq:roundT}
\end{equation}
where $\Delta_i^{(t)}$ is defined by Eq. \ref{eq:avg_lc}. For a specific client $i$, the \emph{accumulated} local model update $\Delta_i$ is:
\begin{equation}
\Delta_i = \sum_{t=0}^{T-1}\Delta_i^{(t)} =  \sum_{t=0}^{T-1}\sum_{j=1}^{k}\beta_i^{(t,j)}g_i(\bm{w}_i^{(t,j)}).
\label{eq:acculc}
\end{equation}

In contrast, for one-shot FL with $T = 1$ , the accumulated local model update is: 
\begin{equation}
\Delta_i = \sum_{j=1}^{Tk}\beta_i^{(0,j)}g_i(\bm{w}_i^{(0,j)}).
\label{eq:onelc}
\end{equation}

Here we set the number of steps per client to $Tk$ since we are trying to match the total number of steps with the multi-round FL.
The reason why the one-shot FL performs worse than the multi-round FL in small models lies in the difference between the local model updates in Eq. \ref{eq:acculc} and Eq. \ref{eq:onelc}. 
In Eq. \ref{eq:acculc}, after the $t$-th communication round, the local training starts from the \emph{updated} global model $\bm{w}^{(t,0)}$, which is aggregated by all the local models in $t$-th round and contains richer knowledge. Therefore, the client can compute a more accurate gradient $g_i(\bm{w}_i^{(t,j)})$ based on the updated model. On the contrary, in one-shot FL (Eq. \ref{eq:onelc}), clients can only continuously train the local models \emph{without} global information. The poor performance of one-shot FL is due to the gradients calculated on the local models being less accurate than those calculated on the aggregated global model. This local error can be expressed in mathematical form:
\begin{equation}
\varepsilon_i = \sum_{j=k+1}^{Tk}\beta_i^{(0,j)}[(g_i(\bm{w}_i^{(0,j)})-g_i(\bm{w}_i^{(t,j-kt)})], \quad \text{where} \ t=\lceil \frac{j}{k} \rceil,
\label{eq:error}
\end{equation}
where $\lceil \cdot \rceil$ means ceiling. $\varepsilon_i$ is the accumulated local update difference between one-shot FL and multi-round FL. 
Since the global model is aggregated by local models, the global model update difference $\varepsilon_i$ is the aggregation of local differences. Its L2 norm can then be bounded by the sum of local differences with triangle inequality, which is:
\begin{equation}
\| \varepsilon \| \leq \sum_{i=1}^{m} \| \varepsilon_i \|.
\label{eq:glerror}
\end{equation}
The global difference can be further simplified by the following assumptions.

\textbf{Assumption 1 (Model Smoothness).} The objective function of the pre-trained large FM is Lipschitz smooth with an $L$ value, that is $\| \nabla F_i(\bm{w}_x) - \nabla F_i(\bm{w}_y) \| \leq L \| \bm{w}_x - \bm{w}_y \|, \ L>0$, where $\nabla F_i(\cdot)$ is the model gradient.

\textbf{Assumption 2 (Bounded Model Updates).} The L2 norm of model updates during FL are much smaller than the initial model parameters, that is, $\| \bm{w}^{(t,j)}-\bm{w}^{(0,0)} \| \leq \tau \|\bm{w}^{(0,0)} \|, \ 0<\tau <1$.

\textbf{Theorem 1 (The one-shot global difference is related to $L$, $\tau$, epoch numbers $Tk$, and number of clients $m$).} Under Assumptions 1 and 2, ignoring the learning rates, the difference between one-shot FL and multi-round FL can be bounded as follows:
\begin{equation}
\| \varepsilon \| \leq \Gamma \|\bm{w}^{(0,0)}\|, \ \text{where} \ \Gamma = L\tau Tkm.
\label{eq:bound}
\end{equation}

This equation indicates that with lower values of $L$, $\tau$, $T$, $k$, and $m$, the model update of one-shot FL will be closer to that of multi-round FL. 
Since our experiments have shown that LLMs exhibit significant advantages over small models in one-shot learning, we conduct experiments on the factors in Equation \ref{eq:bound} to provide a detailed explanation of this phenomenon.

\textbf{Proof of Theorem 1}
According to Eq. \ref{eq:error} and Eq. \ref{eq:glerror}, ignoring the learning rates, the difference of the global model can be bounded by:
\begin{equation}
\varepsilon \leq \sum_{i=1}^{m} \sum_{j=k+1}^{Tk}[(g_i(\bm{w}_i^{(0,j)})-g_i(\bm{w}_i^{(t,j-kt)})],
\label{eq:proof1}
\end{equation}

Considering Assumption 1, we can further factor out the effect of client heterogeneity and write the global model’s cumulative error in an "$m \times \varepsilon_i$" form. Specifically, the aggregate deviation satisfies:
\begin{equation}
\begin{split}
\| \varepsilon \| \leq \sum_{j=k+1}^{Tk} Lm \|(\bm{w}_i^{(0,j)}-\bm{w}_i^{(t,j-kt)}\|,
\end{split}
\label{eq:proof2}
\end{equation}

According to Assumption 2, we can deduce:
\begin{equation}
\begin{split}
\| \varepsilon \| \leq \sum_{j=k+1}^{Tk} L \tau m\|\bm{w}^{(0,0)}\|,
\end{split}
\label{eq:proof3}
\end{equation}

Thus we have:
\begin{equation}
\begin{split}
\| \varepsilon \| \leq L \tau Tk m\|\bm{w}^{(0,0)}\|,
\end{split}
\label{eq:proof4}
\end{equation}
which is Theorem 1. This proof provides the intuition that, smoother model with smaller model updates and less training steps during FL has smaller one-shot global difference, thus performs better in one-shot FL. Motivated by this intuition, we use experimental evidence to show that FMs achieve better one-shot performance in the following paragraphs.

\begin{figure*}[t]
\centering
\hspace{-0.3cm}
\subfloat[The estimated $L$.]{\includegraphics[width=0.32\textwidth]{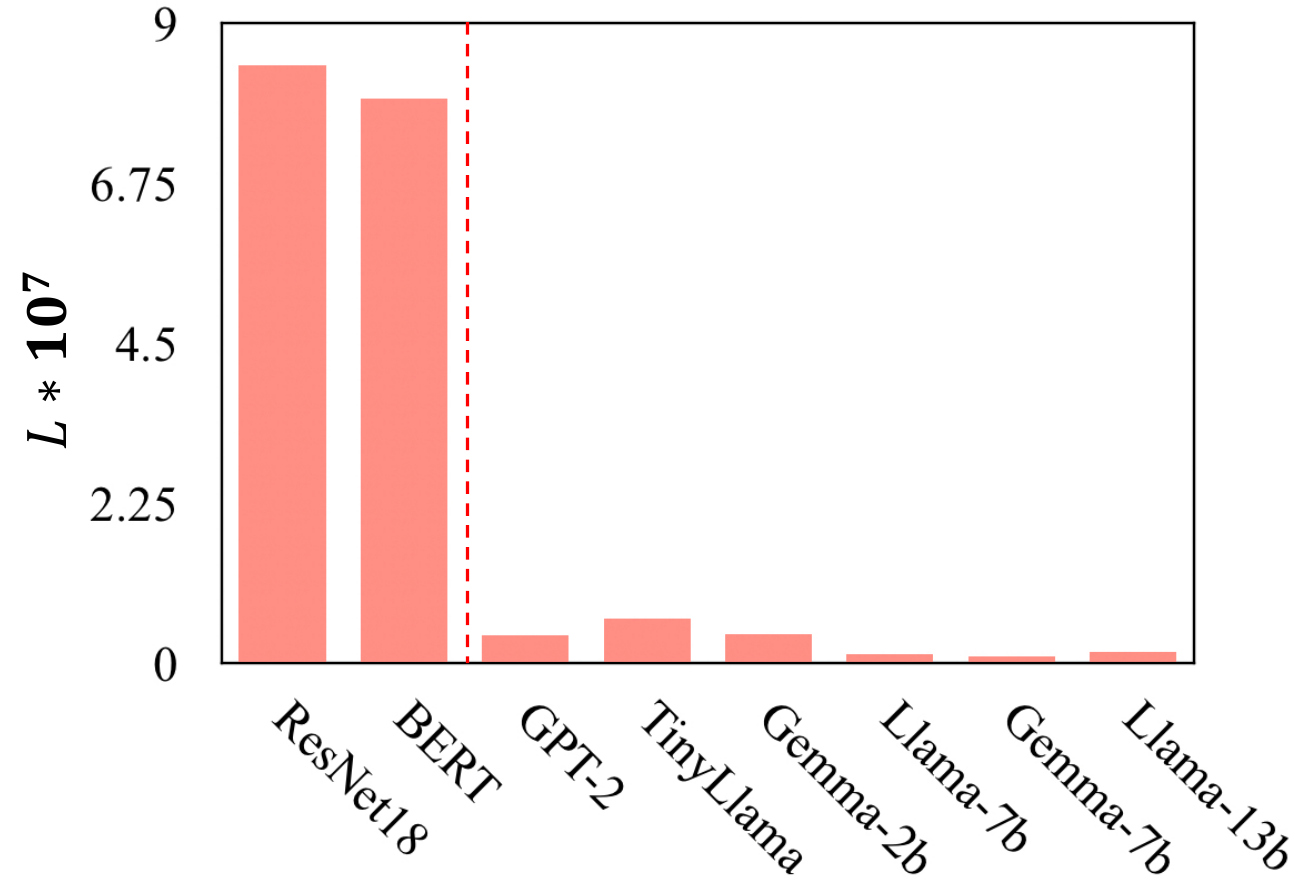}\label{fig:L}}
\hspace{-0.3cm}
\subfloat[The estimated $\tau$.]{\includegraphics[width=0.33\textwidth]{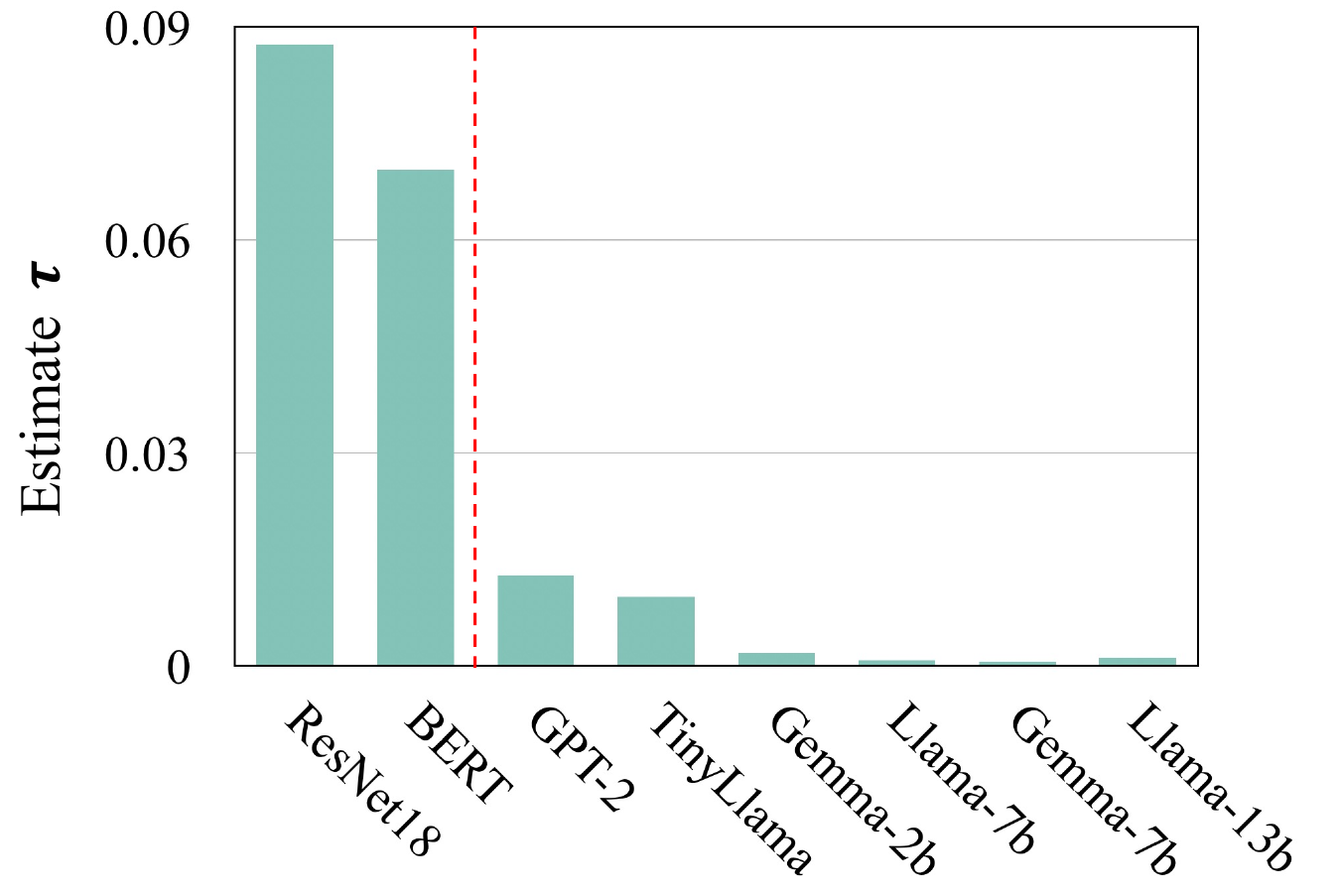}\label{fig:tau}}
\hspace{-0.3cm}
\subfloat[The $\|\bm{w}^{(0,0)} \|$ values.]{\includegraphics[width=0.33\textwidth]{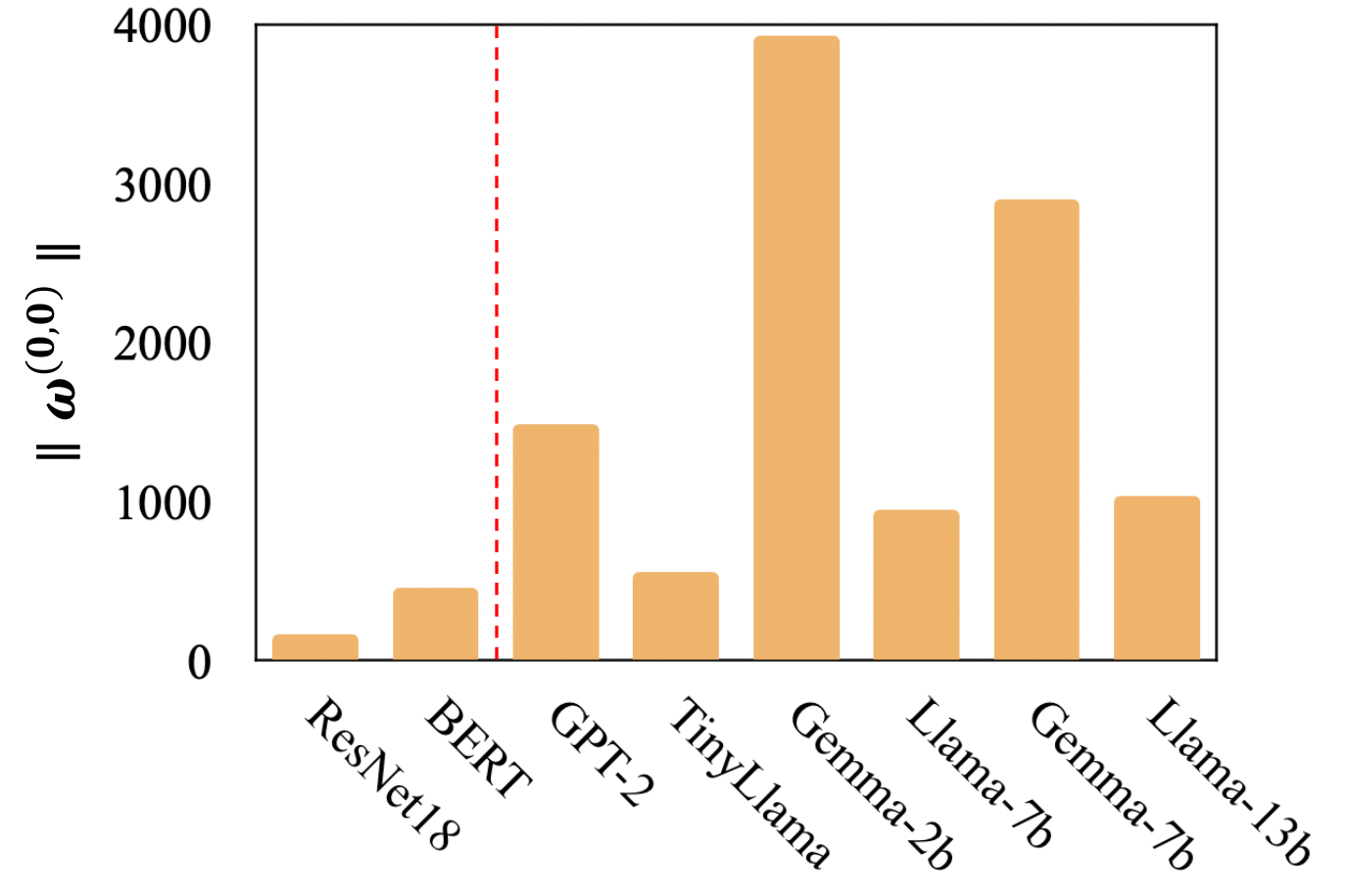}\label{fig:w00}}
\caption{Experiment results on $L$, $\tau$, and $\|\bm{w}^{(0,0)} \|$ in different models. We use the CIFAR-10 dataset to compute the gradient on ResNet18~\cite{he2016deep}. We use the WizardLM dataset~\cite{xu2023wizardlm} to compute the gradient on the language models. 
Models to the left of the red dashed line are small models, while those to the right are foundation models (FMs). The figures indicate that large FMs have significantly smaller $L$ and $\tau$ values compared to small models. 
Additionally, $\|\bm{w}^{(0,0)} \|$ does not increase proportionally with the model size. In conclusion, without considering other unrelated influencing factors, the value of $\Gamma \|\bm{w}^{(0,0)}\|$ decreases as the model size increases.}
\label{fig:bigsmall}
\end{figure*}

\textbf{Foundation Models are Smoother than Small Models ($L_{FM} \ll L_{SM}$).} In Equation \ref{eq:bound}, the factor $L$ represents the smoothness of the model, with smaller $L$ implying a smoother model. We argue that pre-trained large FMs are much smoother than small models and thus have much smaller $L$ values. 
FMs are pre-trained on large-scale datasets to obtain general capabilities. During this pre-training process, the parameters of FMs are optimized from the ridges to the basins in the loss landscape.
Additionally, as observed in a previous work~\cite{ainsworth2022git}, wider models have more flattened basins in the loss landscapes. 
With these pieces of prior knowledge, we hold the contention that the loss landscape in large FM fine-tuning is much \textbf{flatter} and \textbf{smoother} than that in training small models from scratch, resulting in much smaller $L$ values.
To verify this argument, we estimate $L$ by $L = \frac{\| \nabla F_i(\bm{w}_x) - \nabla F_i(\bm{w}_y) \|}{\| \bm{w}_x - \bm{w}_y \|}$. We randomly sample a mini-batch of data in the training datasets and compute the gradient on $\bm{w}^{(0,0)}$ and $\bm{w}^{(T,k)}$ to get $\nabla F_i(\bm{w}^{(0,0)})$ and $\nabla F_i(\bm{w}^{(T,k)})$. Then we visualize the value of $\frac{\| \nabla F_i(\bm{w}^{(0,0)}) - \nabla F_i(\bm{w}^{(T,k)}) \|}{\| \bm{w}^{(0,0)} - \bm{w}^{(T,k)} \|}$ in Fig. \ref{fig:L}. According to Fig. \ref{fig:L}, FMs (\ie models to the right of the red dash line) have much smaller $L$ values than small models, which is consistent with our conjecture.

\textbf{Foundation Models Have Much Smaller Model Updates in Fine-Tuning ($\tau_{FM} \ll \tau_{SM}$).} Another crucial distinction in our analysis lies in the different tasks in FL: \textbf{fine-tuning} and \textbf{training from scratch}. Since the fine-tuning task updates the model parameters to adapt to downstream tasks without compromising its performance on the general task, it will only slightly update the model parameters. Therefore, the model parameter updates in the fine-tuning process are much smaller than the pre-trained model parameters, \ie $\| \bm{w}^{(t,j)}-\bm{w}^{(0,0)} \| \ll \|\bm{w}^{(0,0)} \|$. In this case, the federated fine-tuning task would have a very small $\tau$ in Equation \ref{eq:bound}. To verify this, we conduct experiments to estimate the $\tau$ values by $\frac{\| \bm{w}^{(T,k)}-\bm{w}^{(0,0)} \|}{\|\bm{w}^{(0,0)} \|}$, where $\bm{w}^{(T,k)}$ represents the model update after the entire fine-tuning process on the training datasets. We visualize the estimated $\tau$ values of different models in Fig. \ref{fig:tau}, which illustrates that the $\tau$ values in FMs are much smaller than those in small models.

\begin{figure*}[t]
\centering
\subfloat[One-Shot Aggregation]{\includegraphics[width=0.31\textwidth]{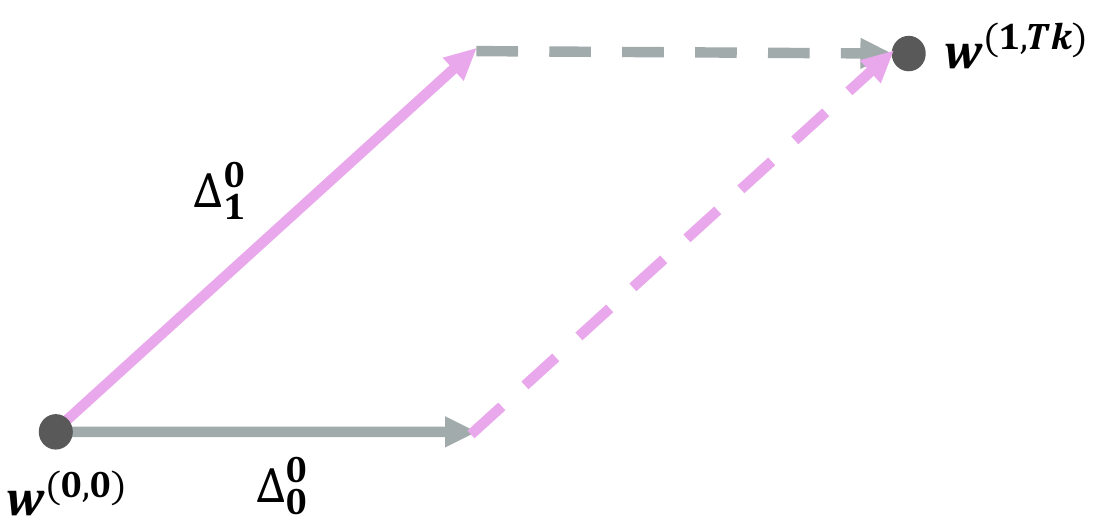}\label{fig:oner}}
\hfil
\subfloat[Multi-Round Aggregation of FMs]{\includegraphics[width=0.33\textwidth]{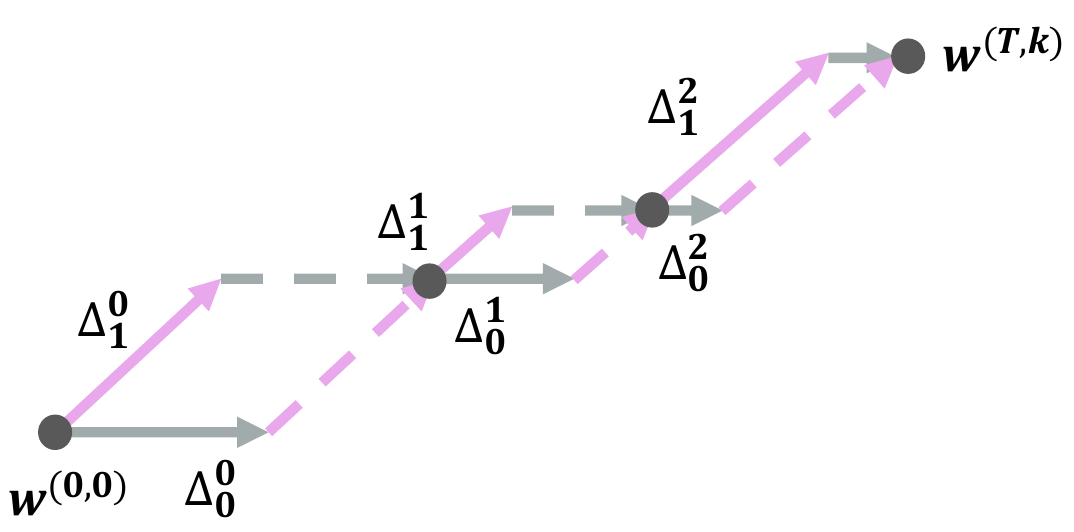}\label{fig:multibase}}
\hfil
\subfloat[Multi-Round Aggregation of SMs]{\includegraphics[width=0.33\textwidth]{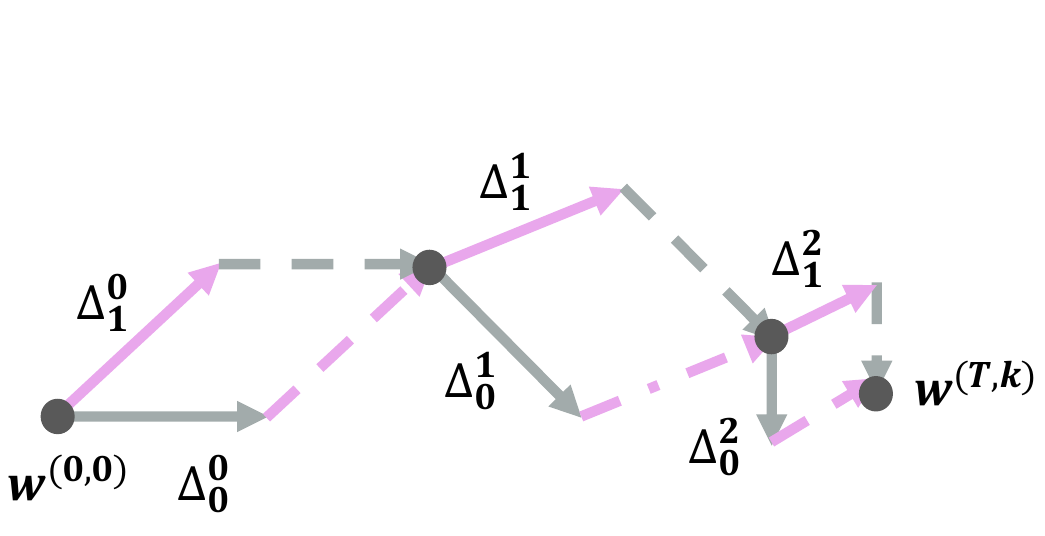}\label{fig:multiours}}
\vspace{0.3cm}
\caption{Aggregation in FL with two clients. (a): one round aggregation with $Tk$ local epochs of both the FM and the small model (SM). Although the FM and the SM may differ in practice, we depict both using the same setup as in (a) to make the comparison easier to understand. (b): FL of FM under $T$ round aggregation with $k$ local epochs, where the landscape is smooth and fine-tuning makes small changes, so the new local directions remain close to the one-shot direction. (c): FL of SM under $T$ round aggregation with $k$ local epochs, where the training direction changes markedly after each global update. The final global model deviates significantly from the one-shot global model. This figure visualizes the intuition behind our analysis that one round works well for FMs.}

\label{fig:intuition}
\end{figure*}


\textbf{Large Foundation Models Require Less Fine-Tuning Steps ($Tk_{FM} \ll Tk_{small}$).} Different from training a small model from scratch, fine-tuning a large model typically doesn't require a large number of total training steps to ensure convergence. This is mainly because the pre-trained models will be overfitting on the fine-tuning data with too many epochs, which will destroy the model's ability on the general tasks. As a result, the $Tk$ values of large FMs are also smaller than those in small models. Table. \ref{tab:tk} displays the $T$ and $k$ numbers adopted by our experiments. 

We also visualize $\| \bm{w}^{(0,0)} \|$ in Fig. \ref{fig:w00}. Although the $\| \bm{w}^{(0,0)} \|$ value of the small model is relatively small, it does not exhibit a clear trend positively correlated with model size (\eg TinyLlama has a similar $\| \bm{w}^{(0,0)} \|$ value with BERT, but has 10 times more parameters than BERT, Gemma-2b has much larger $\| \bm{w}^{(0,0)} \|$ value than Llama-13b).

\begin{figure}[h]
    \centering
    \includegraphics[width=0.5\textwidth]{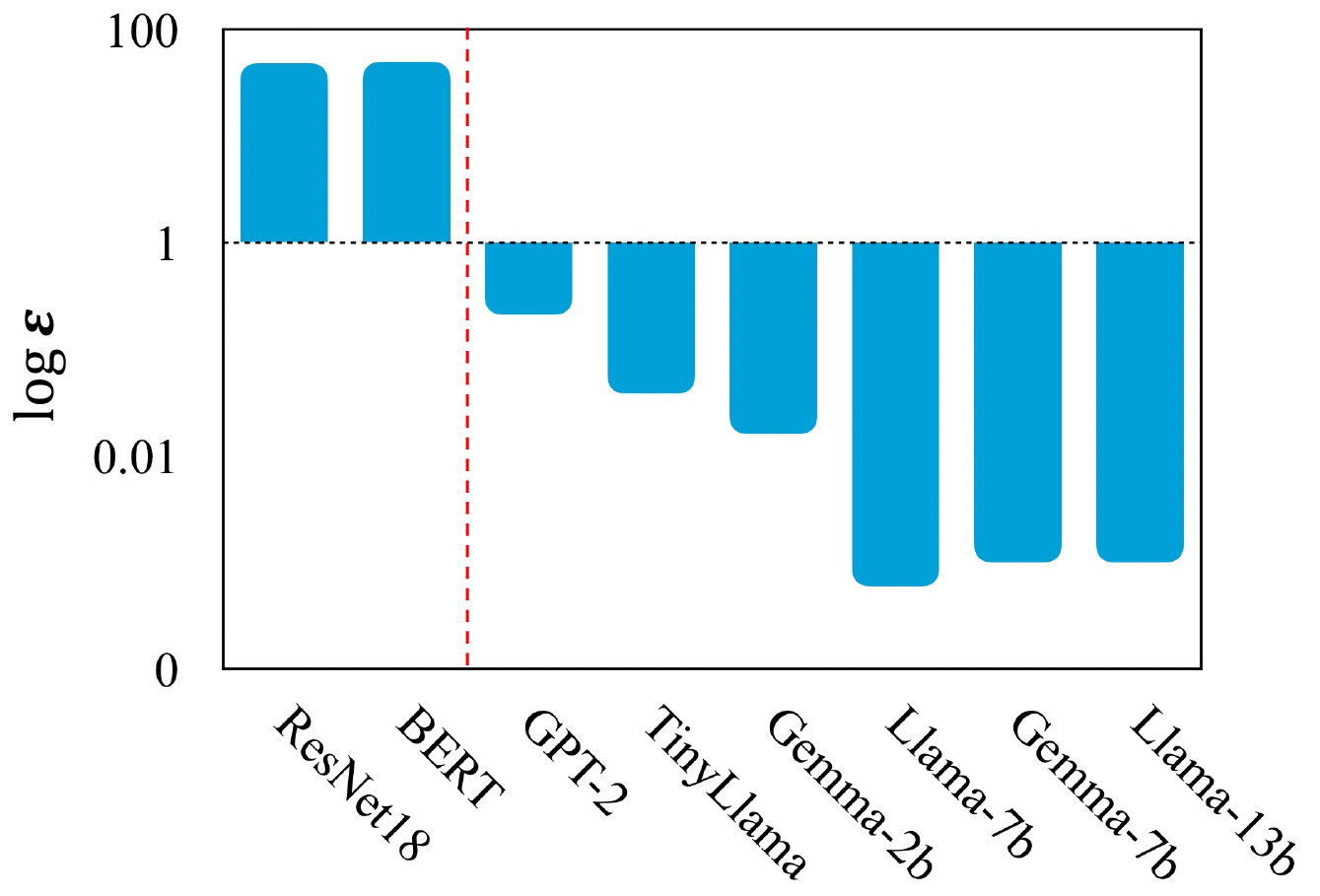}
    \caption{The estimated $\log \| \varepsilon \|$ in different models calculated by $\log \| \varepsilon \| = \log (L\tau Tk \|\bm{w}^{(0,0)} \|)$.}
    \label{fig:final}
\end{figure}

\textbf{Conclusion: Large Foundation Models Have Smaller Global Difference $\varepsilon$.} Based on the discussion before regarding the $L$, $\tau$, $Tk$, and $\|\bm{w}^{(0,0)} \|$ values of the model with various sizes, we conclude that large FMs have smaller $L$, $\tau$, and $Tk$ values, while $\|\bm{w}^{(0,0)} \|$ is not strongly related to the model size. We ignore the client number $m$ and visualize the $\| \varepsilon \| = \Gamma \|\bm{w}^{(0,0)}\|$ values of different models in Fig. \ref{fig:final}. The results in Fig. \ref{fig:final} clearly demonstrate that large FMs (GPT-2 and all models to its right) have significantly lower $\| \varepsilon \|$ values than the small models, with larger FMs having lower values. According to Eq. \ref{eq:bound}, smaller $\| \varepsilon \|$ means a smaller difference between one-shot and multi-round FL. Consequently, FMs have much better one-shot FL performance than small models. 

In summary, the reasons why large FMs have smaller differences in one-shot federated fine-tuning are due to three main factors. \textbf{\emph{First, the pre-trained FMs have extremely smooth loss landscapes in fine-tuning, \ie $L_{FM} \ll L_{SM}$. Second, the fine-tuning model updates are particularly small compared to the pre-trained parameters, \ie $\tau_{FM} \ll \tau_{SM}$. Third, FM fine-tuning requires far fewer epochs than training small models from scratch, \ie $Tk_{FM} \ll Tk_{SM}$. These three factors lead to much smaller error $\varepsilon$ in the one-shot federated fine-tuning of FMs.}}

Our theoretical analysis focuses on the error gap between one-shot and standard multi-round FL, namely the discrepancy caused by using a single communication round, rather than on asymptotic convergence. Unlike classical FL results that prove convergence as $T\!\to\!\infty$~\cite{wang2020tackling,kairouz2021advances}, Theorem~1 expresses this gap as an explicit function of model- and task-dependent quantities (e.g., smoothness, update magnitude, and training horizon), thereby predicting how \emph{model scale} and \emph{task regime} (fine-tuning v.s. training from scratch) influence one-shot FL. These predictions are borne out in our experiments, whereas asymptotic analyses do not capture such finite-round, size- and task-sensitive effects.

These proofs and experiments also support the following intuition: because the loss landscape of large pre-trained models is relatively smooth and fine-tuning induces only small parameter changes, the update directions remain largely stable even when the clients’ starting points shift across rounds in multi-round FL. As a result, when we switch to one-shot FL and do not refresh the starting point, the effective direction remains close to that of multi-round FL, which yields an outcome similar to vector aggregation. In contrast, smaller models encounter a more rugged loss landscape, so clients require multiple FL rounds to repeatedly adjust the update direction and search for a better optimum. Under one-shot FL, each client continues to fine-tune from the initial parameters throughout, which causes error accumulation and ultimately degrades performance. Fig.~\ref{fig:intuition} describes this intuition. The error of one-shot FL in small model accumulates with the multi-round global aggregation (Fig.~\ref{fig:intuition} (c)), while the error of FMs (Fig.~\ref{fig:intuition} (b)) remains small due to the smooth loss landscape and small fine-tuning updates.

\section{Experiment}
\label{sec:exp}
\subsection{Experimental Setups}
\textbf{Models and Datasets.} To demonstrate the performance of FMs of different sizes on one-shot federated fine-tuning, we selected multiple models ranging in parameter size from 1b to 13b for experiments. The language models we experimented with range in parameter size from smallest to largest as follows: TinyLlama (1.1b)~\cite{zhang2024tinyllama}, Gemma-2b, Gemma-7b~\cite{team2024gemma}, Llama-7b, and Llama-13b~\cite{touvron2023llama}. We use the MMLU~\cite{hendrycks2020measuring} training dataset and Wizard~\cite{luo2023wizardmath} dataset to federated fine-tune these models. For evaluation, we leverage MMLU and ARC Challenge~\cite{allenai:arc} in Eval-Harness~\cite{eval-harness} to evaluate the model ability of QA tasks, and the GPT-4 evaluation in MT-bench~\cite{zheng2023judging} for the chat assistant task.

\begin{table*}[t] 
\caption{$Tk$ settings in experiments. $T$ is the number of global communication rounds. $k$ is the total number of local SGD steps, which is computed by (dataset length $\times$ epoch number $/$ batch size).}
	\label{tab:tk}
	\centering
	\footnotesize
        \resizebox{18cm}{!}{
	\begin{tabular}{c|cc|cccccc}
		\toprule
 &ResNet-18&BERT &\textbf{GPT-2}
&\textbf{TinyLlama}&\textbf{Gemma-2b}&\textbf{Llama-7b}&\textbf{Gemma-7b}&\textbf{Llama-13b} \\
        \midrule
        \textbf{T}&50&50&5&3&3&3&3&3\\
        \midrule
        \textbf{k}&7812&3906&5625&3750&1875&1875&1875&1875\\
        \midrule
        \textbf{Tk}&390600&195300&28125&11250&5625&5625&5625&5625\\
		\bottomrule
	\end{tabular}}
\end{table*}

\begin{table*}[t]
  \caption{Performance of multi-round and one-shot federated fine-tuning in Q\&A tasks. The rows with star (*) are the results of one-shot federated fine-tuning.%
  }
  \label{exp:qa}
  \centering
  \resizebox{18.0cm}{!}{
   \begin{tabular}{cc|cccccccccccc}
    \toprule
    \multirow{2}{*}{\textbf{Tasks}} &\multirow{2}{*}{\textbf{Methods}} &\multicolumn{3}{c}{\textbf{TinyLlama}} &\multicolumn{3}{c}{\textbf{Gemma-2b}}&\multicolumn{3}{c}{\textbf{Llama-7b}}&\multicolumn{3}{c}{\textbf{Llama-13b}}\\
    \cmidrule(r){3-5}
    \cmidrule(r){6-8}
    \cmidrule(r){9-11}
    \cmidrule(r){12-14}
    &&\textbf{MMLU}&\textbf{Wizard}&\textbf{M-W}&\textbf{MMLU}&\textbf{Wizard}&\textbf{M-W}&\textbf{MMLU}&\textbf{Wizard}&\textbf{M-W}&\textbf{MMLU}&\textbf{Wizard}&\textbf{M-W}\\
    \midrule
    \multirow{4}{*}{\textbf{MMLU}} &LoRA &\textbf{25.08} &\textbf{25.07} &24.98 &\textbf{38.43} &\textbf{37.75} &\textbf{37.69} &\textbf{36.16} &35.07 &\textbf{35.37} &47.22 &46.83 &46.82 \\
    &\textbf{LoRA*} &25.01 &25.04 &\textbf{25.03} &38.24 &36.55 &35.14 &35.86 &\textbf{35.91} &34.84 &\textbf{48.40} &\textbf{47.93} &\textbf{47.43} \\
    &Full FT &\textbf{27.30} &24.84 &\textbf{25.46} &\textbf{42.02} &\textbf{34.60} &28.36 &\textbf{45.61} &30.52 &28.81 &\textbf{50.24} &\textbf{42.12} &\textbf{32.91} \\
    &\textbf{Full FT*} &26.39 &\textbf{24.87} &24.99  &40.93 &33.86 &\textbf{28.71} &44.20 &\textbf{33.97} &\textbf{29.05} &48.30 &39.62 &29.76 \\
    \midrule
    \multirow{4}{*}{\textbf{ARC}} &LoRA &35.49 &\textbf{37.28} &\textbf{36.69} &\textbf{43.09} &\textbf{43.26} &42.06 &50.43 &50.94 &51.19 &55.72 &55.72 &55.63  \\
    &\textbf{LoRA*} &\textbf{36.86} &36.77 &36.26 &40.61 &42.49 &\textbf{42.15} &\textbf{50.85} &\textbf{51.88} &\textbf{52.13} &\textbf{56.40} &\textbf{58.11} &\textbf{56.74}  \\
    &Full FT &32.76 &\textbf{37.03} &33.02 &\textbf{41.04} &45.48 &\textbf{37.46} &\textbf{43.26} &40.24 &\textbf{37.15} &42.41 &\textbf{47.57} &\textbf{42.75}  \\
    &\textbf{Full FT*} &\textbf{33.19} &36.26 &\textbf{33.87} &39.85 &\textbf{45.92} &34.47 &41.72 &\textbf{43.52} &37.03 &\textbf{44.62} &45.05 &40.21  \\
    \bottomrule
  \end{tabular}}
\end{table*}

\begin{table}[t]
\centering
\caption{Performance of one-shot federated fine-tuning on chat assistant tasks. Wizard has better performance than MMLU on MT-bench. We use AVG. column to show the averaging performance of specific methods.}
\resizebox{0.48\textwidth}{!}{
    \begin{tabular}{cc|ccc|c|c}
        \toprule
        \textbf{Models} &\textbf{Methods} &\textbf{MMLU}&\textbf{Wizard} &\textbf{M-W}&\textbf{AVG.}&\textbf{Base}\\
        \midrule
        \multirow{4}{*}{\textbf{TinyLlama}} &LoRA &3.59 &3.44 &3.65 &\textbf{3.56} &\multirow{4}{*}{3.47}  \\
        &\textbf{LoRA*} &3.33 &3.45 &3.74 &3.51  \\
        &Full FT &2.02 &3.76 &2.97 &\textbf{2.92}  \\
        &\textbf{Full FT*} &1.91 &4.21 &2.38 &2.83  \\
        \midrule
        \multirow{4}{*}{\textbf{Gemma-2b}} &LoRA &3.36 &3.48 &3.46 &3.43 &\multirow{4}{*}{3.60}  \\
        &\textbf{LoRA*} &3.23 &3.77 &3.66 &\textbf{3.55}  \\
        &Full FT &2.16 &4.36 &2.75 &\textbf{3.09}  \\
        &\textbf{Full FT*} &1.92 &4.27 &2.50 &2.90 \\
        \midrule
        \multirow{4}{*}{\textbf{Llama-7b}} &LoRA &3.01 &3.27 &2.99 &3.09 &\multirow{4}{*}{2.86} \\
        &\textbf{LoRA*} &2.69 &3.90 &3.54 &\textbf{3.38}  \\
        &Full FT &1.85 &4.18 &2.31 &2.78  \\
        &\textbf{Full FT*} &1.56 &4.79 &2.21 &\textbf{2.85} \\
        \midrule
        \multirow{4}{*}{\textbf{Llama-13b}} &LoRA &2.58 &2.68 &2.86 &2.71 &\multirow{4}{*}{2.69}   \\
        &\textbf{LoRA*} &3.02 &4.27 &3.26 &\textbf{3.52} \\
        &Full FT &2.43 &4.63 &3.05 &\textbf{3.37} \\
        &\textbf{Full FT*} &1.81 &4.74 &2.62 &3.06 \\
        \bottomrule
    \end{tabular}
}
\label{fig:chat_table}
\end{table}

\textbf{Federated Fine-Tuning Settings.} For federated fine-tuning on a single MMLU or Wizard dataset, we randomly split the dataset into 10 clients. We also have a strongly non-iid setting, which assigns the MMLU dataset to 10 clients and the Wizard dataset to another 10 clients, and lets the 20 clients fine-tune the FM. For the baseline, we use a multi-round FedAvg algorithm on both LoRA and full fine-tuning. For our one-shot federated fine-tuning, we only perform a single communication round. To ensure fairness, we keep the total number of local steps the same between multi-round and one-shot federated fine-tuning. \eg, if the setting in multi-round federated fine-tuning is 3 communication rounds, 1 local epoch in each round, the setting in one-shot should be 1 communication round, 3 local epochs in that round.
For LoRA fine-tuning across all the models and datasets, we set the local LoRA rank to 16, the local learning rate to 3e-4, and the batch size to 64. For full fine-tuning, we reduced the learning rate to 3e-5 and set the batch size to 8. For multi-round settings, the numbers of global communication rounds and local epochs in each round in different models and datasets are listed in Table \ref{exp:settings}. The one-shot setting satisfies $T=1$ and $k$ equals $Tk$ in the multi-round setting. The number of rounds and epochs we selected can ensure convergence and avoid overfitting. We show a simple example in Fig. \ref{fig:multi} to demonstrate this point. Note that, given the inherent heterogeneity of text datasets (\eg differing sequence lengths), heterogeneity is present in all our experiments.

\textbf{Computer Resources.}For LoRA fine-tuning on all the models and full fine-tuning on all the models except Llama-13b, we used 4 NVIDIA RTX A6000 GPUs. For Llama-13b full fine-tuning, we use 8 NVIDIA A100 GPUs.

\begin{table*}[h] 
\caption{Global rounds and local epochs settings in multi-round experiments.}
	\label{exp:settings}
	\centering
	\footnotesize
        \resizebox{18cm}{!}{
	\begin{tabular}{c|cccccccccccc}
		\toprule
\textbf{Models}&\multicolumn{3}{c}{\textbf{TinyLlama}}&\multicolumn{3}{c}{\textbf{Gemma-2b}}&\multicolumn{3}{c}{\textbf{Llama-7b}}&\multicolumn{3}{c}{\textbf{Llama-13b}}\\
&MMLU&Wizard&M-W&MMLU&Wizard&M-W&MMLU&Wizard&M-W&MMLU&Wizard&M-W\\
\midrule
\textbf{T}&3&3&3&3&3&3&3&3&3&3&3&3\\
\textbf{k}&1&2&1&1&2&1&2&1&1&1&1&1\\
		\bottomrule
	\end{tabular}}
\end{table*}

\subsection{Main Results}
\textbf{One-Shot Federated Fine-Tuning in QA Tasks.} 
We first evaluate the performance of one-shot federated fine-tuning in QA tasks and display the results in Table \ref{exp:qa}. The columns with titles MMLU, Wizard, and M-W represent the model fine-tuned by MMLU, Wizard, and the mixture of MMLU and Wizard datasets respectively. The rows with the title MMLU and ARC represent the model accuracy evaluated by the MMLU test set and ARC Challenge. The Methods columns mean the fine-tuning is performed by LoRA or full fine-tuning, while the rows with a star (*) represent one-shot federated fine-tuning.
According to Table \ref{exp:qa}, the performance of one-shot federated fine-tuning is generally comparable to that of multi-round federated fine-tuning. In some settings, one-shot fine-tuning even achieves higher accuracy. For example, the Llama-13b model one-shot fine-tuned by LoRA on the Wizard dataset achieves 47.93\% accuracy on MMLU and 58.11\% on ARC Challenge, which is higher than the 46.83\% and 55.72\% accuracy of multi-round fine-tuning. 
In full fine-tuning, multi-round fine-tuning performs better in some settings. For instance, the Llama-13b model multi-round full fine-tuned on the Wizard dataset outperforms one-shot fine-tuning on both MMLU and ARC Challenge. 
These observations align with our previous theoretical analysis. 
Full fine-tuning involves greater parameter updates compared to LoRA, resulting in a larger $\tau$ value, and thus a larger $\varepsilon$ value. Consequently, the performance of one-shot full fine-tuning may sometimes be inferior to LoRA fine-tuning. However, this does not affect our overall conclusion: \textbf{for large FMs, one-shot federated fine-tuning can effectively replace multi-round federated fine-tuning.} One-shot fine-tuning provides comparable performance to multi-round fine-tuning while significantly reducing communication costs. To clarify the fine-tuning gain, we test the zero-shot performance of models used in Table \ref{exp:qa} for reference. The results are displayed in Table \ref{tab:zero}.

\textbf{One-Shot Federated Fine-Tuning in Chat Assistant Tasks.}
We evaluate the performance of FMs in chat assistant tasks, where models generate answers to several questions and are scored by GPT-4. The score from MT-bench is the average score across all questions.
Table \ref{fig:chat_table} shows the scores of multi-round and one-shot federated fine-tuned models. 
The averaging scores of three fine-tuning datasets indicate that larger FMs perform better in one-shot federated fine-tuning. 
Specifically, multi-round fine-tuning outperforms one-shot fine-tuning in both LoRA and full fine-tuning on the TinyLlama model, which is the smallest model in our experiments. On the contrary, for larger models, such as Gemma-7b and Llama-13b, one-shot fine-tuning performs better than multi-round fine-tuning. This observation aligns with our previous theoretical analysis that larger models have smaller differences between one-shot and multi-round FL. 
The superior performance of one-shot fine-tuning in larger models might be attributed to the less number of local epochs per round, which leads to a slower local learning rate decay. The chat assistant's capabilities may benefit from this smoother learning rate decay process.

\begin{figure}[t]
\centering
\includegraphics[width=0.95\linewidth]{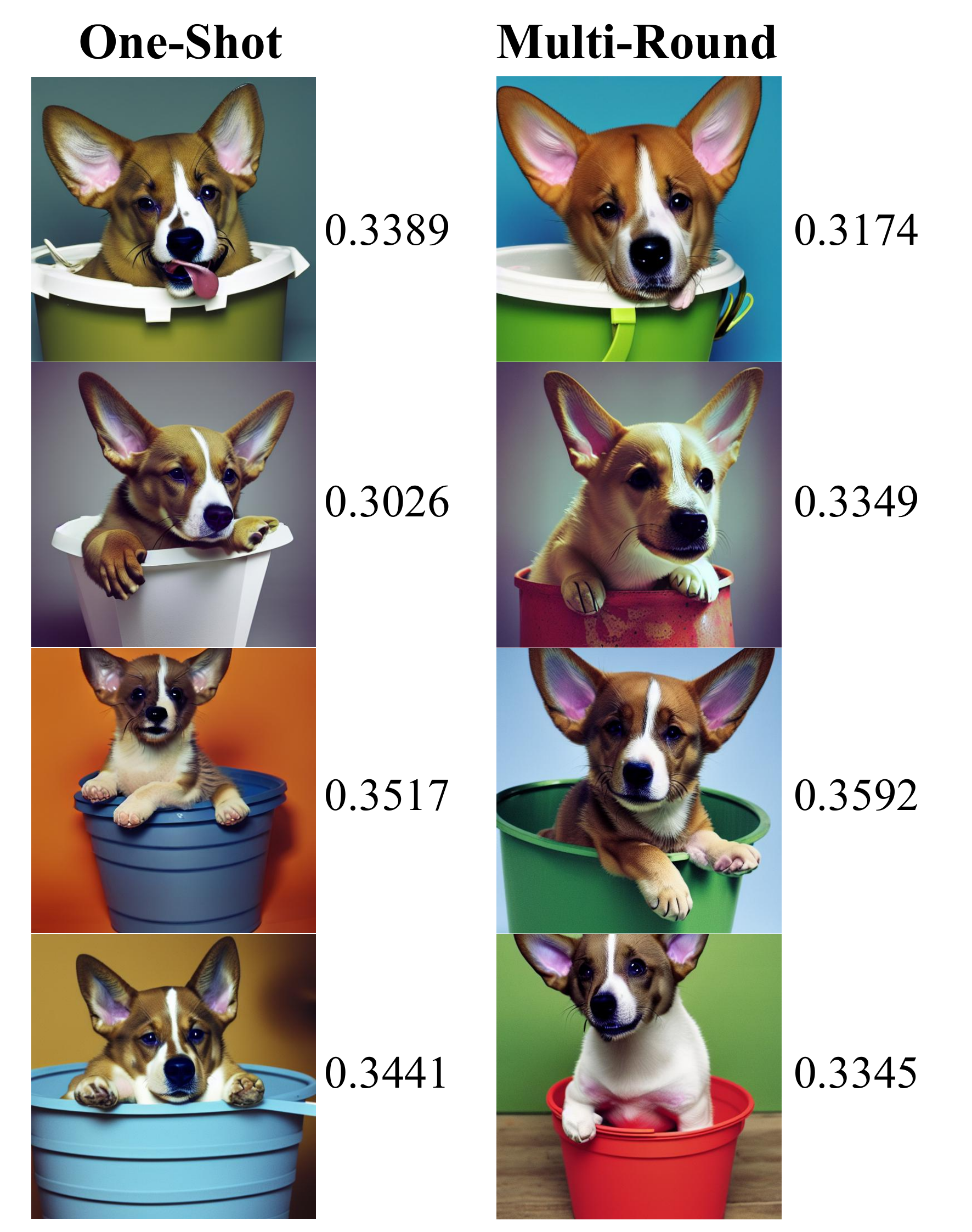}
\caption{``A photo of a dog in a bucket'' generated by LoRA fine-tuned stable diffusion models. The left column shows images generated by the one-shot FL models, along with their CLIP scores; the right column shows the results of multi-round FL.}
\label{fig:chat_demo}
\end{figure}

\textbf{One-Shot Federated Fine-Tuning in Text-To-Image Generation Tasks.}
In addition to testing LLMs, we also evaluated the effectiveness of one-shot federated fine-tuning in the text-to-image generation tasks. We use LoRA to fine-tune a stable-diffusion-v1-5~\cite{rombach2022high} model on the Dreambooth~\cite{ruiz2023dreambooth} dataset with 5 distributed clients. In the multi-round setting, we have 5 global rounds, with 5 local epochs in each round. In the one-shot setting, we have 1 global round and 25 local epochs in that round.
After fine-tuning, we evaluated the models using the CLIP~\cite{hessel2021clipscore} score with ViT-B-32~\cite{dosovitskiy2020image} to assess the quality of generated images based on specific prompts. Fig. \ref{fig:chat_demo} shows the images generated with the prompt "A photo of a dog in a bucket"
The right column displays the result of multi-round federated fine-tuning, while the left column shows the result from the one-shot setting. 
The numbers to the right of the images represent the CLIP scores. 
The qualities of the images generated by both methods are essentially the same. The average CLIP score in the one-shot setting is 0.3343, while the score in the multi-round setting is 0.3345. These results indicate that the effectiveness of one-shot federated fine-tuning extends to fine-tuning stable diffusion models.

\begin{table}[t] 
\caption{Zero-Shot results of models on MMLU and ARC Challenge.}
	\label{tab:zero}
	\centering
	\footnotesize
        \resizebox{3.4in}{!}{
	\begin{tabular}{c|cccc}
		\toprule
\textbf{Tasks}&\textbf{TinyLlama}&\textbf{Gemma-2b}&\textbf{Llama-7b}&\textbf{Llama-13b} \\
        \midrule
        \textbf{MMLU}&24.90&34.63&34.44&46.23\\
        \textbf{ARC}&35.41&40.25&45.65&51.79\\
		\bottomrule
	\end{tabular}}
\end{table}


\begin{figure*}[t]
\centering
\includegraphics[width=\textwidth,height=0.21\textwidth,keepaspectratio=false]{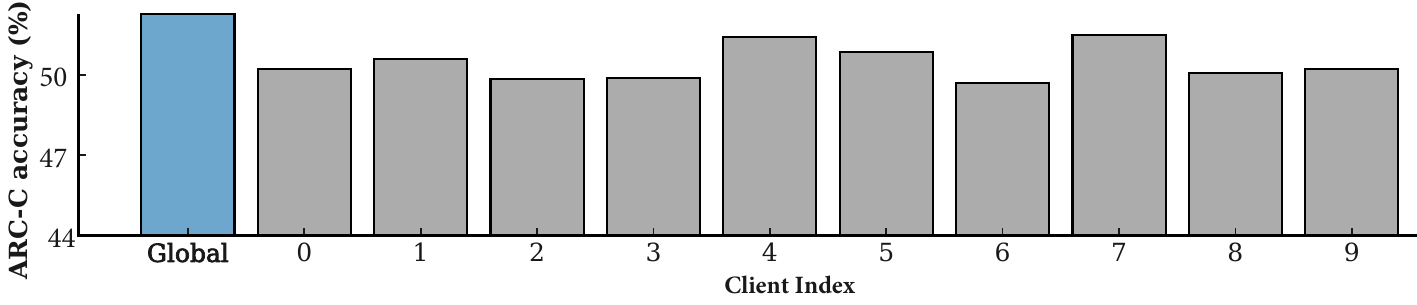}
\caption{Standalone results of one-shot federated fine-tuning on Llama-7B with Wizard dataset. Bars 0–9 are the accuracy of local models with the corresponding index; \textbf{Global} is the one-shot aggregated global model.}
\label{fig:standalone}
\end{figure*}

\textbf{Standalone Results of Local Models.} To further demonstrate the effectiveness of one-shot federated fine-tuning, we performed the standalone experiment to compare the performance of the global model and the local model only trained on local datasets. We did the experiments on the llama-7b model and Wizard dataset and displayed the results in Fig. \ref{fig:standalone}. The results show that the accuracy of local models is slightly lower than that of the global model, with some local models perform similar to the global model. This is reasonable in the context of the federated fine-tuning task because the models have already been pre-trained. Therefore, even though clients have less training data, the performance of local models does not differ significantly from the global model. This further supports that a single aggregation can already capture most of the attainable gain.

\begin{figure}[t]
    \centering
    \includegraphics[width=0.49\textwidth]{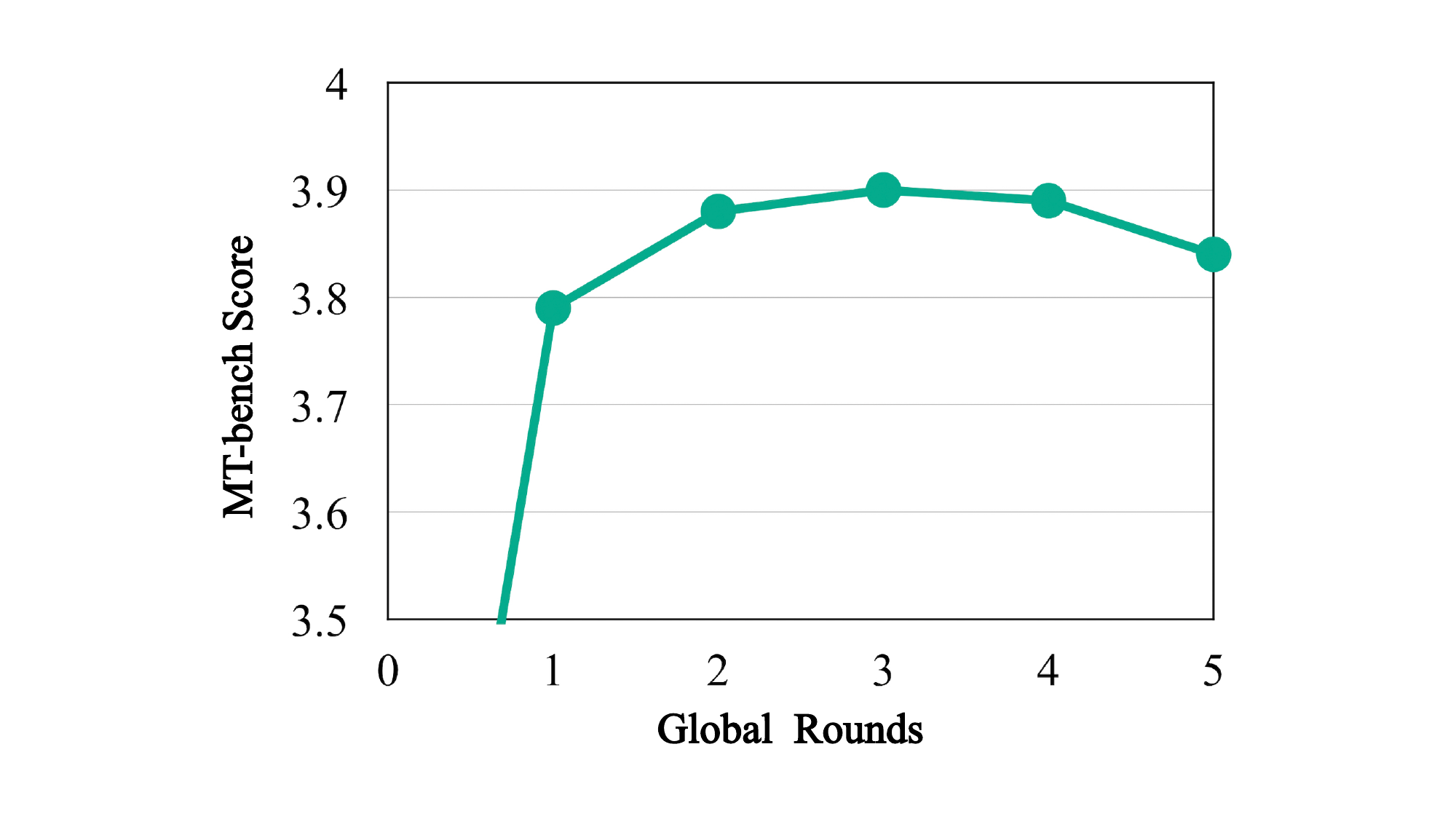}
    \caption{The MT-bench score of the global model in 1–5 global rounds.}
    \label{fig:multi}
\end{figure}

\textbf{More Global Round Settings.} We also tested the model performance when we had more and fewer global rounds in a multi-round setting. We evaluated the global model in 1, 2, 3, 4, and 5 global rounds when fine-tuning the Llama-7b model on Wizard dataset. The results are shown in Fig. \ref{fig:multi}. In the first round, the MT-bench score increases from the 2.86 in base model to around 3.80. Then, it slightly increases towards 3.90 in the 3rd round and begins to decrease afterward. A similar phenomenon can be seen in other datasets and models that the model performance will increase in the initial 2-4 rounds and then gradually decline due to overfitting. Thus, we use 3 global rounds in all of the multi-round experiments.

\begin{figure}[t]
    \includegraphics[width=0.5\textwidth]{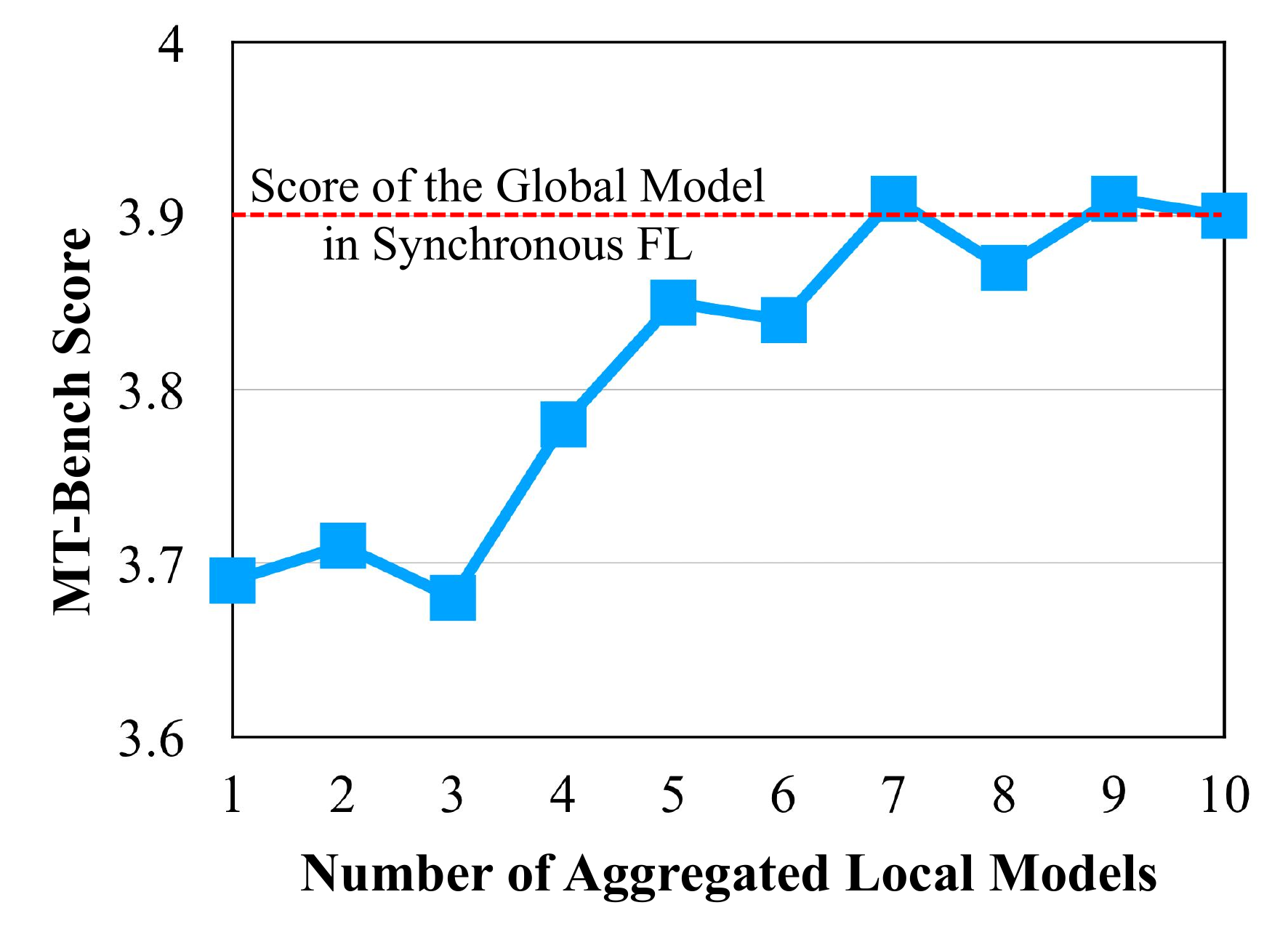}
    \caption{The MT-bench score of the global model merged by a varied number of clients.}
    \label{fig:clients}
\end{figure}

\section{Discussion}
\label{sec:dis}

\paragraph{One-Shot Federated Fine-Tuning Saves Communication Cost}
In FL, the server needs to send the model parameters to all the selected clients and receive the model updates from the clients in each communication round~\cite{ghiasvand2024communication}. Thus, the total number of communicated parameters in multi-round should be $2mTS$, where $S$ is the model size. In one-shot federated fine-tuning, the server and the clients only perform one-round communication, so the number of communicated parameters is only $2mS$. This reduction in communication overhead is significant, especially in the federated fine-tuning of large FMs. For instance, the Llama-13b model has approximately 50\,GB parameters, \ie $S=50$\,GB. In our experiments, the three-round federated fine-tuning on Llama-13b needs to communicate about 3000\,GB data between the server and the clients, which may be unaffordable in scenarios with tight communication budgets, whereas one-shot reduces this amount to about 1000\,GB. Beyond raw transfer size, fewer rounds simplify orchestration and reduce exposure to transient failures during long jobs, since there are fewer synchronization points and fewer opportunities for timeouts. In practice, one-shot can be combined with standard techniques such as quantization, sparsification, and parameter-efficient tuning so that the effective $S$ is smaller, which further improves communication efficiency without changing the algorithmic interface. This makes the approach easier to deploy in bandwidth-constrained environments and lowers the barrier for cross-organization collaboration where network conditions and maintenance windows are variable.

\paragraph{One-Shot Federated Fine-Tuning Supports Asynchronous Global Aggregation}
In traditional multi-round FL, clients need to train local models synchronously. The server can only perform the aggregation and send the new global model to clients after receiving all local model updates. This requirement poses challenges for federated learning applications. For example, if local computation resources are occupied by other tasks or if the connection between the server and clients is unstable, the training process will be halted. One-shot federated fine-tuning effectively addresses this problem. The server can update the global model with local updates as soon as they are received, allowing for real-time model updates. Therefore, even if some clients fail to send model updates promptly due to various reasons, the global model on the server can still be updated by most clients, resulting in a usable global model. To further illustrate this point, we sequentially aggregated local model updates from client 1 to client 10 in one-shot federated fine-tuning of Llama-7b on the Wizard dataset. We tested the global model's performance on the MT-bench as we aggregated updates from 1, 2, 3, and up to 10 clients. The results are displayed in Fig.~\ref{fig:clients}. The model score increases as more clients contribute their local updates to the global model, indicating that each individual local model update provides an immediate improvement in global model performance. The red dashed line represents the model score in the synchronous FL setting, which matches the score of aggregating ten clients in the asynchronous setting. This behavior enables flexible participation under heterogeneous compute and partial availability, reduces server idle time, and better utilizes short or sporadic connectivity, which is often the case for edge devices and inter-organization deployments. From a systems perspective, it also simplifies recovery from stragglers and node restarts, since the aggregation state can progress whenever any subset of clients completes.

\paragraph{One-Shot Federated Fine-Tuning Naturally Mitigates Client-Side Privacy Threats}
In a traditional FL algorithm, clients repeatedly receive new global model parameters each round, which could lead to client-side privacy issues. Malicious clients can exploit model inversion~\cite{fredrikson2015model,zhang2020secret} and gradient inversion attacks~\cite{huang2021evaluating} to recover private training samples or user inputs from other clients~\cite{wei2023client}. These attacks heavily rely on access to the global model parameters and certain data distribution information. However, in one-shot FL, the server can choose not to send back global parameters to the clients and only provide an API of the fine-tuned model. By doing this, it can eliminate the possibility of client-side privacy attacks. This design shifts the trust boundary from repeated parameter release to a service interface with authentication, access control, and logging, which is easier to monitor and audit. It is also compatible with defenses such as secure aggregation and client-side differential privacy, which are orthogonal to the one-shot protocol and can be enabled without changing the communication pattern. We note that while this reduces a class of client-side threats, it does not by itself address server-side risks such as model extraction; appropriate rate limiting, watermarking, and usage policies remain necessary in deployments that expose a public or partner-facing API.

\section{Related Work}
\paragraph{One-Shot Federated Learning} One-shot federated learning refers to learning the parameters of the global model in a single round of communication between clients and the server~\cite{guha2019one,liu2025one,guan2025capture,amato2025towards}. There are two main strategies for optimizing one-shot FL, neuron matching and knowledge distillation. Neuron matching is based on the permutation symmetry of neural networks~\cite{ainsworth2022git}, which means that client model parameters can be aligned according to a common ordering and then be averaged. Previous works use algorithms such as the Fisher information matrix~\cite{jhunjhunwala2024fedfisher} and permutation matrix~\cite{wang2020federated} to match the local model parameters.
The knowledge distillation methods aim at distilling knowledge from well-trained local models through public data~\cite{gong2021ensemble,li2020practical,heinbaugh2022data}. Some works also use distilled data to transfer knowledge between clients and the server~\cite{zhou2020distilled}. Recent works adopt generative models to help generate substitute data for the local dataset on the server~\cite{yang2024exploring, zhang2022dense}. These lines of work primarily seek to improve the protocol or optimization of one-shot FL (often for small or medium models), whereas we empirically \emph{discover} and theoretically \emph{explain} that one-shot FL already performs well for FMs. Hence we do not provide exhaustive head-to-head comparisons with every algorithmic variant; our contribution is to document the phenomenon and to offer an error analysis, rather than to introduce a new one-shot algorithm. Importantly, our findings are compatible with these techniques: neuron matching or distillation can be plugged into our setting, and our theory clarifies when and why such one-shot procedures should be effective in FMs.

\paragraph{Federated Fine-Tuning} Federated fine-tuning~\cite{cheng2021fine,orescanin2021federated} aims to adapt foundation models (FMs) on cross-domain, on-device datasets while preserving data privacy. To reduce communication and computation, many works employ PEFT methods such as LoRA~\cite{hu2021lora} within the federated setting~\cite{zhang2024towards}. As in classical FL, federated fine-tuning must contend with non-IID data~\cite{wang2024flora, cho2024heterogeneous} and personalization~\cite{wagner2024personalized}. In practice, the main deployment bottlenecks are the computational and bandwidth constraints of edge devices~\cite{woisetschlager2024federated,wang2025mobilea3gent,zheng2023autofed}, for which one-shot FL can alleviate part of the cost.

While most federated fine-tuning studies on FMs pragmatically adopt \emph{fewer communication rounds} and \emph{shorter local training} for efficiency, prior work has not explicitly posed the \emph{difference between FMs and small models} as a research question. Our work fills this gap: we empirically \emph{identify} and theoretically \emph{explain} why one-shot FL tends to perform well specifically for FMs, thereby offering a new lens on federated fine-tuning. Moreover, our findings are closely related to \emph{task arithmetic}, \emph{task vectors}, and \emph{model merging}~\cite{hendel2023context,ilharco2022editing,zhao2024texttt}, lines of work that have largely flourished in the FM era. This coincidence suggests a unifying hypothesis: the same factors that make one-shot FL effective for FMs (\eg smoother landscapes, smaller update magnitudes, shorter effective horizons) may also underlie the empirical success of model merging and task arithmetic. These observations motivate a broader research agenda that explains and predicts the divergent behaviors of FMs versus small models in federated fine-tuning.

\section{Conclusion}
In this paper, we tackle the critical issue of high communication costs that limit the practical application of federated fine-tuning. Through a series of experiments, we demonstrate that multi-round communication is not necessary for fine-tuning FMs, as one-shot federated fine-tuning achieves comparable performance. We then provide a theoretical analysis to explain why one-shot federated fine-tuning is effective for large FMs and validate our findings with empirical evidence.
Our extensive experiments show that one-shot federated fine-tuning performs on par with multi-round federated fine-tuning across 5 different models and 3 diverse tasks. This method significantly reduces communication overhead, making federated fine-tuning more feasible and efficient, especially for large-scale models. Moreover, one-shot federated fine-tuning supports asynchronous local updates and enhances security by minimizing data exposure during the training process.

Although this work does not propose a new algorithm, it offers a principled explanation for why one-shot FL can work well in large models. The analysis clarifies the conditions under which the one-shot gap remains small and provides insight into related phenomena such as task vectors and model editing, where small, targeted updates also transfer effectively in large models. Importantly, the empirical results align with the theoretical predictions and show that the bounds and assumptions connect with practical training behavior rather than remaining purely abstract.
These findings make it possible to harness the power of large FMs in environments with limited communication resources, thereby broadening the accessibility and utility of advanced AI technologies. We view this as an initial step toward a more theory-grounded understanding of communication-efficient adaptation in federated settings, and we expect the framework to inform future studies on robust aggregation in FL under realistic system constraints.

\bibliographystyle{IEEEtran}
\bibliography{reference}












\begin{IEEEbiography}[{\includegraphics[width=1in,height=1.25in,clip,keepaspectratio]{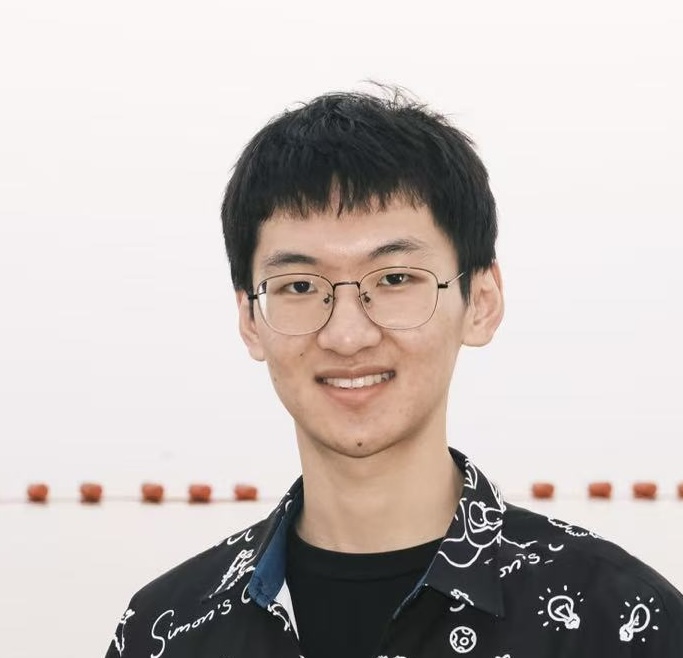}}]{}
\textbf{Ziyao Wang} is a third-year Ph.D. student in the Department of Electrical and Computer Engineering at the University of Maryland, College Park, advised by Prof. Ang Li. He received his B.E. degree from Wuhan University, China. His research focuses on improving the efficiency and trustworthiness of collaborative LLM systems, with interests in areas such as federated learning and server-device collaboration. He is also broadly interested in machine learning security and privacy. His research have been published in top conferences and journals such as NeurIPS, ICML, ICLR, EMNLP, CCS, TIFS, and TDSC. He has served as a reviewer for top-tier conferences (NeurIPS, ICML, ICLR, EMNLP) and journals (TMC, IEEE IoT).
\end{IEEEbiography}

\begin{IEEEbiography}[{\includegraphics[width=1in,height=1.25in,clip,keepaspectratio]{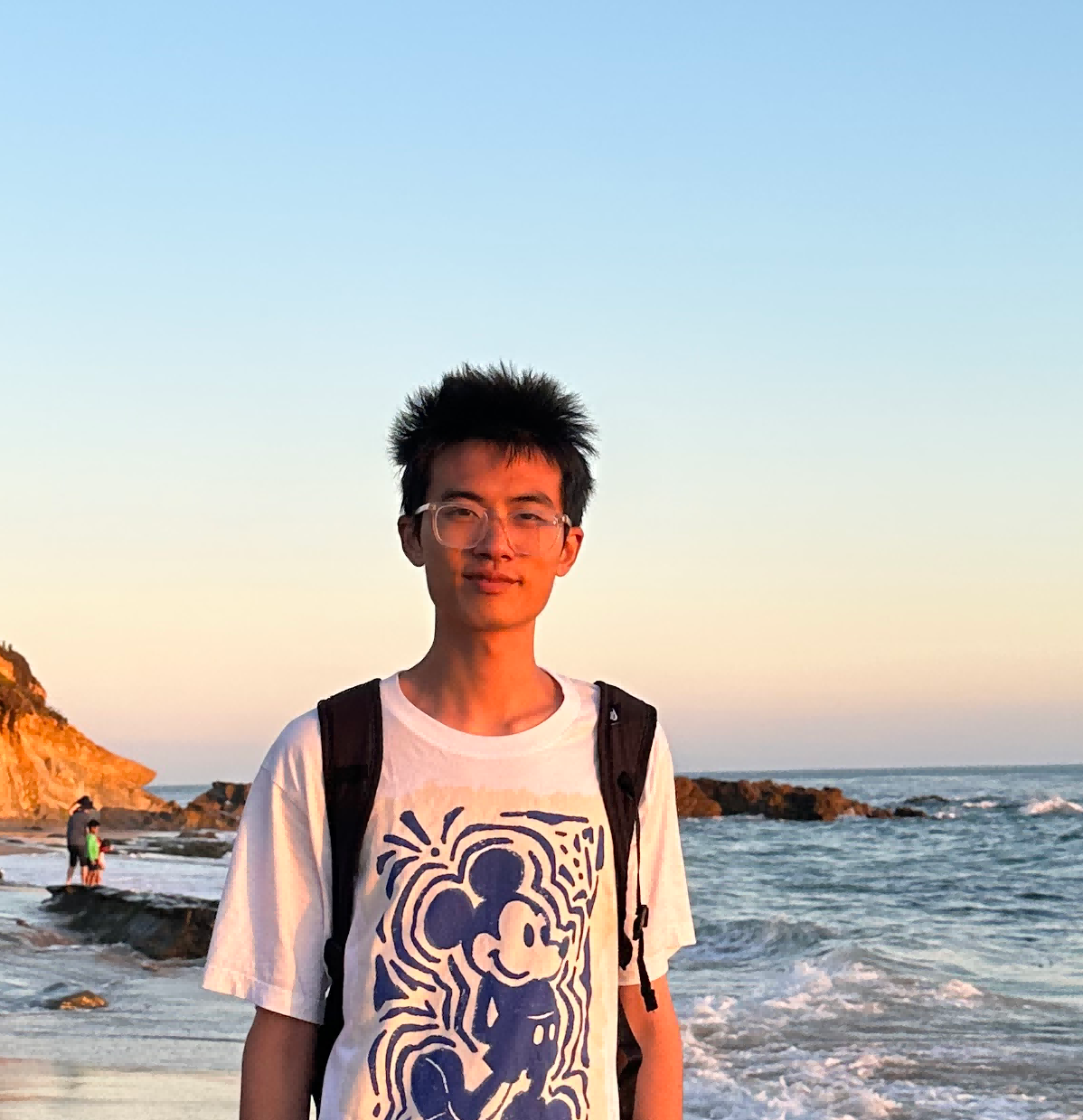}}]{}
Bowei Tian is a second-year Ph.D. student in the Department of Electrical and Computer Engineering at the University of Maryland, College Park, advised by Prof. Ang Li. He received his B.E. degree from Wuhan University, China. His research interests include causal reasoning and representation learning, AI security and privacy, and computer vision. His work has been published in top conferences and journals such as NeurIPS, ICLR, EMNLP, Ubicomp, ECCV, and TDSC. He has also served as a reviewer for top-tier conferences, including NeurIPS and AAAI.
\end{IEEEbiography}

\begin{IEEEbiography}[{\includegraphics[width=1in,height=1.25in,clip,keepaspectratio]{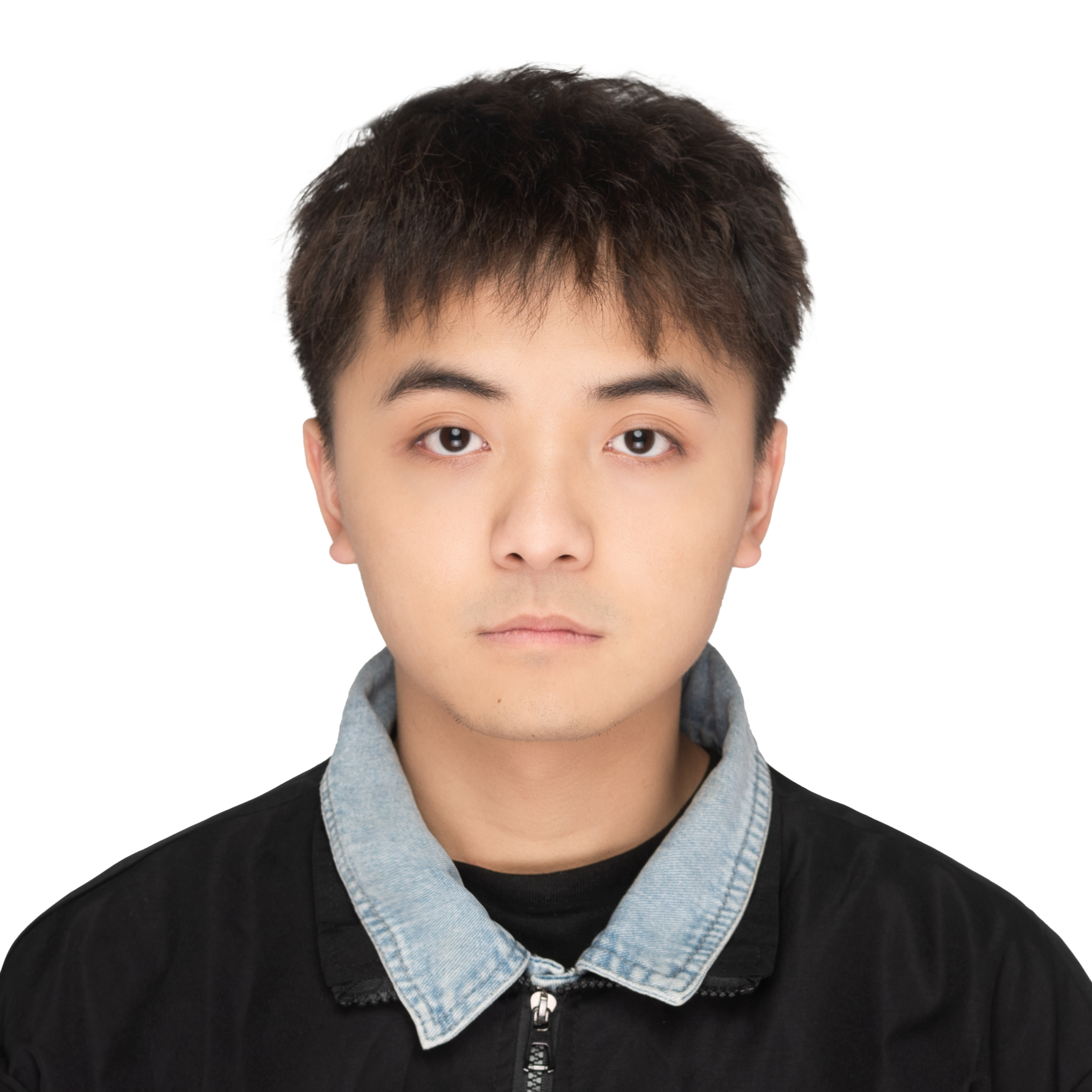}}]{}
Yexiao He is a third-year Ph.D. student in the Department of Electrical and Computer Engineering at the University of Maryland, College Park, advised by Prof. Ang Li. He received his B.E. and M.E. degrees from the University of Electronic Science and Technology of China. His research focuses on LLMs, AI for healthcare, and neuro-symbolic AI. His work has been published in top conferences such as NeurIPS and MobiSys.
\end{IEEEbiography}

\begin{IEEEbiography}[{\includegraphics[width=1in,height=1.25in,clip,keepaspectratio]{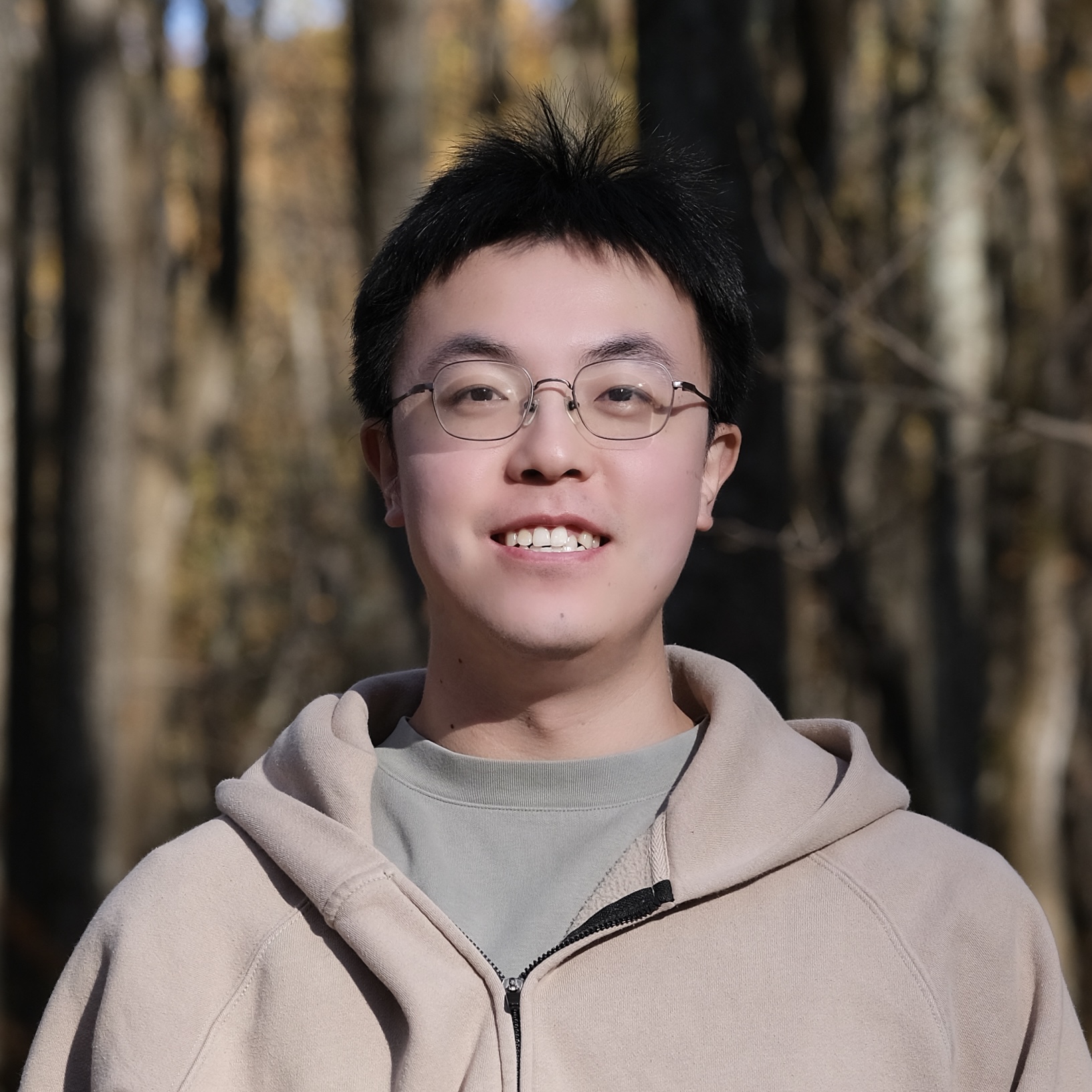}}]{}
Zheyu Shen is a third-year Ph.D. student in the Department of Electrical and Computer Engineering at the University of Maryland, College Park, advised by Prof. Ang Li. He received his M.S. degree in Computer Science from the University of Southern California and his B.E. degree from Northwestern Polytechnical University, China. His research focuses on machine learning systems, with particular interests in efficient LLM training and serving, federated learning, and distributed systems. He is also broadly interested in topics at the intersection of machine learning and edge computing. His work has been published at top conferences such as NeurIPS, EMNLP and MobiSys. He has also served as a reviewer for ICLR and NeurIPS.
\end{IEEEbiography}

\begin{IEEEbiography}[{\includegraphics[width=1in,height=1.25in,clip,keepaspectratio]{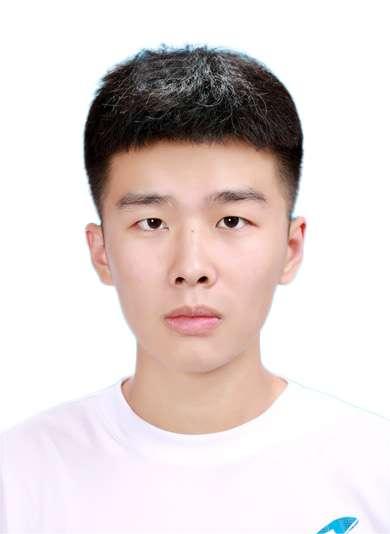}}]{}
Guoheng Sun is a second-year Ph.D. student in the Department of Electrical and Computer Engineering at the University of Maryland, College Park, advised by Prof. Ang Li. He received his B.E. degree from Sichuan University, China. His research interests lie in improving the efficiency and trustworthiness of LLMs, particularly through training-free or training-less approaches. He is also broadly interested in privacy, safety alignment, and the pretraining process of LLMs. He has serves(d) as a reviewer for top-tier conferences such as NeurIPS and ICLR. He is one of the awardees of the 2025 North America Qualcomm Innovation Fellowship.
\end{IEEEbiography}

\begin{IEEEbiography}[{\includegraphics[width=1in,height=1.25in,clip,keepaspectratio]{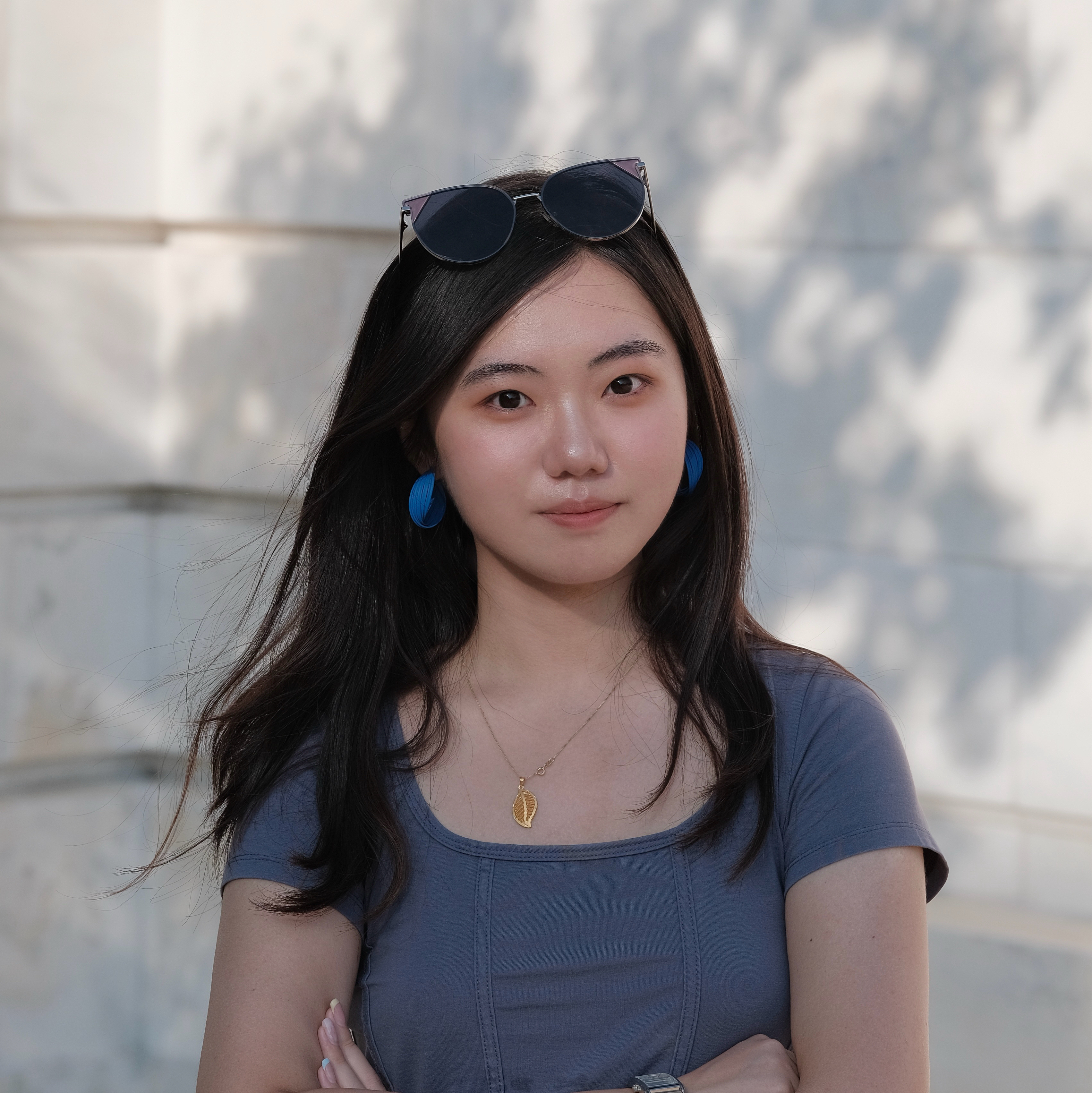}}]{}
Yuhan Liu is a first-year Ph.D. student in the School of Electronic Engineering and Computer Science at Queen Mary University of London, affiliated with the Centre for Digital Music. She received her Bachelor’s and Master’s degrees from Beijing Institute of Technology, China. Her research interests focus on machine learning for music analysis, specifically music source separation and music understanding.
\end{IEEEbiography}

\begin{IEEEbiography}[{\includegraphics[width=1in,height=1.25in,clip,keepaspectratio]{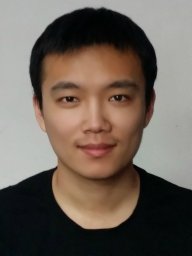}}]{}
Luyang Liu is a Research Scientist at Google DeepMind, focusing on foundation models, representation learning, and federated learning. He is a core contributor to Gemini models, with a focus on its thinking capability. He is also interested in enabling LLM with new capabilities, including latent thinking and text diffusions (i.e., Gemini Diffusion). Before that, he worked on representation learning, federated learning, and health research at Google Research. Before joining Google, Luyang conducted research in mobile/edge computing and road safety during his Ph.D. at Rutgers. His works have been published in top-tier venues such as ICML, NeurIPS, AAAI, MobiCom, MobiSys, Nature, and Science.
\end{IEEEbiography}

\begin{IEEEbiography}[{\includegraphics[width=1in,height=1.25in,clip,keepaspectratio]{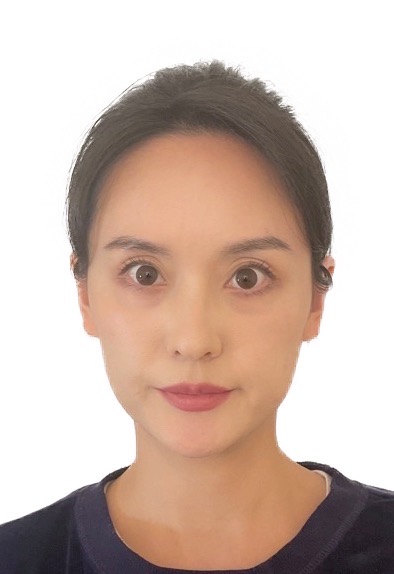}}]{}
\textbf{Meng Liu} received her M.S. degree from University of Maryland, College Park, and B.S. from Henan University, China. Her research interests focus on interpretable machine learning, efficient machine learning, and AI for financial data.
\end{IEEEbiography}

\begin{IEEEbiography}[{\includegraphics[width=1in,height=1.25in,clip,keepaspectratio]{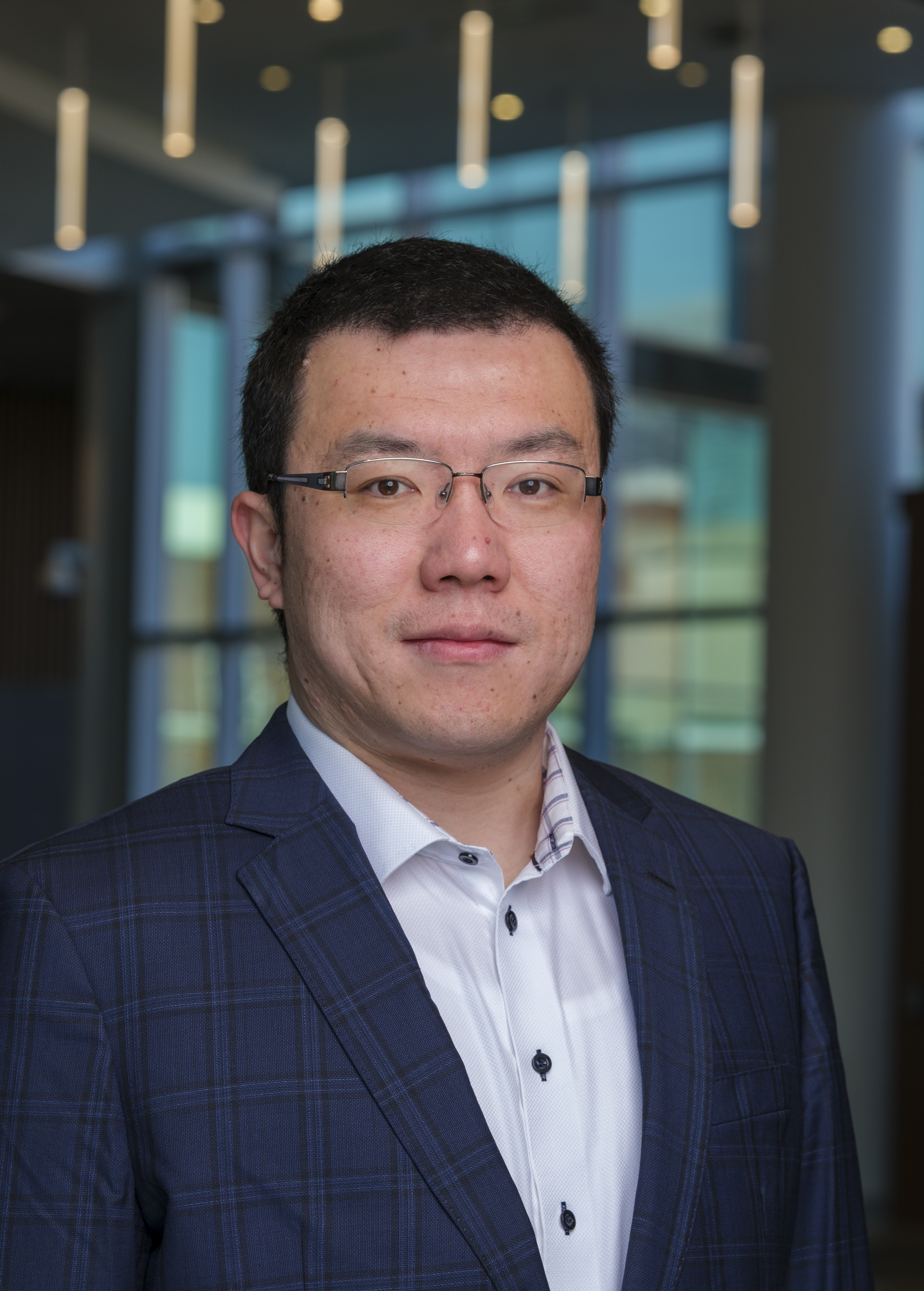}}]{}
\textbf{Ang Li} is an Assistant professor in the Department of Electrical and Computer Engineering at the University of Maryland College Park. Before joining UMD, he was a research associate at Qualcomm AI Research. He received Ph.D. in Electrical and Computer Engineering from Duke University in 2022. His research interests lie in the intersection of machine learning and edge computing, with a focus on building large-scale networked and efficient intelligent systems. His research has been recognized with a variety of awards, including the 2024 ACM CCS Distinguished Paper Award, 2022 IEEE TCCPS Outstanding Ph.D. Dissertation Award, 2022 Duke ECE Department Outstanding Dissertation Award, and ACM KDD Best Student Paper Award in 2020.
\end{IEEEbiography}

\end{document}